\documentclass{sigkddExp}

\usepackage{enumitem}
\usepackage{xcolor}
\DeclareMathOperator*{\argmax}{argmax}
\DeclareMathOperator*{\argmin}{argmin}

\begin{document}
%

\title{Adversarial Attacks and Defenses:\\ An Interpretation Perspective}
%

\numberofauthors{1}
%


\author{
%
\alignauthor Ninghao Liu$^{\dagger}$, Mengnan Du$^{\dagger}$, Ruocheng Guo$^{\ddagger}$, Huan Liu$^{\ddagger}$, Xia Hu$^{\dagger}$ \\
      \affaddr{Department of Computer Science and Engineering, Texas A\&M University, TX, USA}\\
      \affaddr{Computer Science \& Engineering, Arizona State University, Tempe, AZ, USA}\\
      \email{$^{\dagger}$\{nhliu43, dumengnan, xiahu\}@tamu.edu, $^{\ddagger}$\{rguo12, huanliu\}@asu.edu}
}

\date{30 July 1999}
\maketitle
\begin{abstract}
Despite the recent advances in a wide spectrum of applications, machine learning models, especially deep neural networks, have been shown to be vulnerable to \textit{adversarial attacks}. Attackers add carefully-crafted perturbations to input, where the perturbations are almost imperceptible to humans, but can cause models to make wrong predictions. Techniques to protect models against adversarial input are called \textit{adversarial defense} methods. Although many approaches have been proposed to study adversarial attacks and defenses in different scenarios, an intriguing and crucial challenge remains that how to really understand model vulnerability? Inspired by the saying that ``if you know yourself and your enemy, you need not fear the battles", we may tackle the aforementioned challenge after interpreting machine learning models to open the black-boxes. The goal of \textit{model interpretation}, or \textit{interpretable machine learning}, is to extract human-understandable terms for the working mechanism of models. Recently, some approaches start incorporating interpretation into the exploration of adversarial attacks and defenses. Meanwhile, we also observe that many existing methods of adversarial attacks and defenses, although not explicitly claimed, can be understood from the perspective of interpretation.
In this paper, we review recent work on adversarial attacks and defenses, particularly from the perspective of machine learning interpretation. We categorize interpretation into two types, feature-level interpretation and model-level interpretation. For each type of interpretation, we elaborate on how it could be used for adversarial attacks and defenses. We then briefly illustrate additional correlations between interpretation and adversaries. Finally, we discuss the challenges and future directions along tackling adversary issues with interpretation.
\end{abstract}

\section*{Keywords}
Adversarial attack, model robustness, interpretation, explainability, deep learning

\section{Introduction}
Machine learning (ML) techniques, especially recent deep learning models, are progressing rapidly and have been increasingly applied in various applications. Nevertheless, concerns have been posed about the security and reliability issues of ML models. In particular, many deep models are susceptible to adversarial attacks~\cite{Good-etal15explaining, Szeg-etal13intriguing}. That is, after adding certain well-designed but human imperceptible perturbation or transformation to a clean data instance, we are able to manipulate the prediction of the model. The data instances after being attacked are called \textit{adversarial samples}. The phenomenon is intriguing since clean samples and adversarial samples are usually not distinguishable to a human. Adversarial samples may be predicted dramatically differently from clean samples, but the predictions usually do not make sense to a human.

\begin{figure}[t]
\centering
 \vspace{0pt}
 \includegraphics[width=0.38\textwidth]{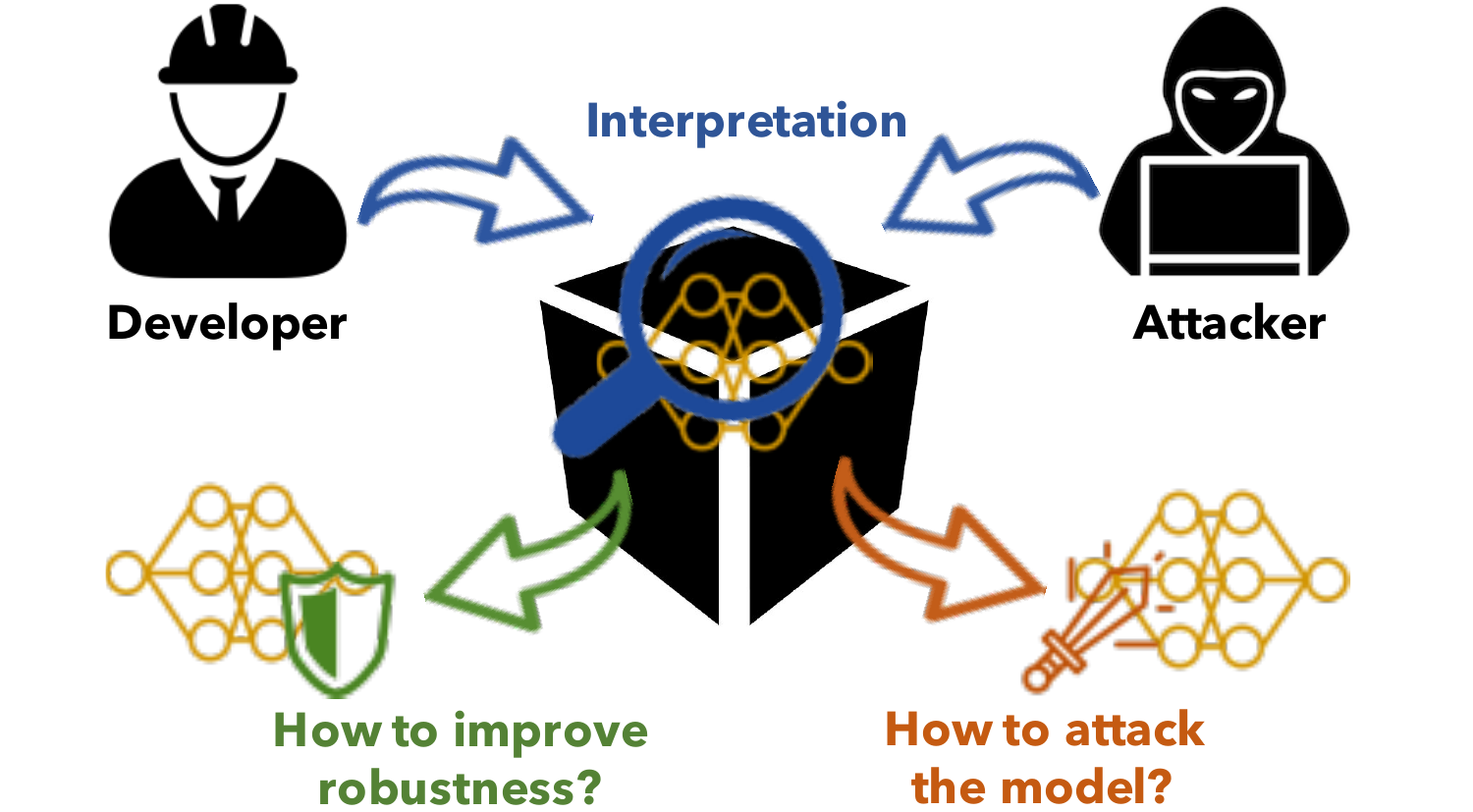}
 \vspace{0pt}
 \caption{Interpretation can either provide directions for improving model robustness or attacking on its weakness. } \label{fig:highlevel}
\end{figure}

The model vulnerability to adversarial attacks has been discovered in various applications or under different constraints. For examples, approaches for crafting adversarial samples have been proposed in tasks such as classification (e.g., on image data~\cite{Kurakin-etal17atScale}, text data~\cite{Lei-etal19discreteText}, tabular data~\cite{Liu-etal18adversarial}, graph data~\cite{zugner2018adversarial}), object detection~\cite{song2018physical}, and fraud detection~\cite{zeager2017adversarial}. Adversarial attacks could be initiated under different constraints, such as assuming limited knowledge of attackers on target models~\cite{Pape-etal16practical, Pape-etal16transferability}, assuming higher generalization level of attack~\cite{Moosavi-etal17universalPerturb, Mopuri-etal17universalDataIndependent}, posing different real-world constraints on attack~\cite{Thys-Ranst19surveilanceCamera, Kurakin-etal16physical}. Given these progresses, several questions could be posted. First, are these progresses relatively independent of each other, or is there a underlying perspective from which we are able to discover the commonality behind them? Second, should adversarial samples be seen as the negligent corner cases that could be fixed by putting patches to models, or are they deeply rooted to the internal working mechanism of models that it is not easy get rid of?

Motivated by the idiom that ``if you know yourself and your enemy, you need not fear the battles" from \textit{The Art of War}, in this paper, we answer the above questions and review the recent advances of adversarial attack and defense approaches from the perspective of interpretable machine learning. The relation between model interpretation and model robustness is illustrated in Figure~\ref{fig:highlevel}. On one hand, if adversaries know how the target model work, they may utilize it to find model weakness and initiate attacks accordingly. On the other hand, if model developers know how the model works, they could identify the vulnerability and work on remediation in advance. Interpretation refers to the human-understandable information explaining what a model have learned or how a model makes prediction. Exploration of model interpretability has attracted many interests in recent years, because recent machine learning techniques, especially deep learning models, have been criticized due to lack of transparency. Some recent work starts to involve interpretability into the analysis of adversarial robustness. Also, although not being explicitly specified, in this survey we will show that many existing adversary-related work can be comprehended from another perspective as extension of model interpretation.

\begin{figure}[t]
\centering
 \vspace{0pt}
 \includegraphics[width=0.48\textwidth]{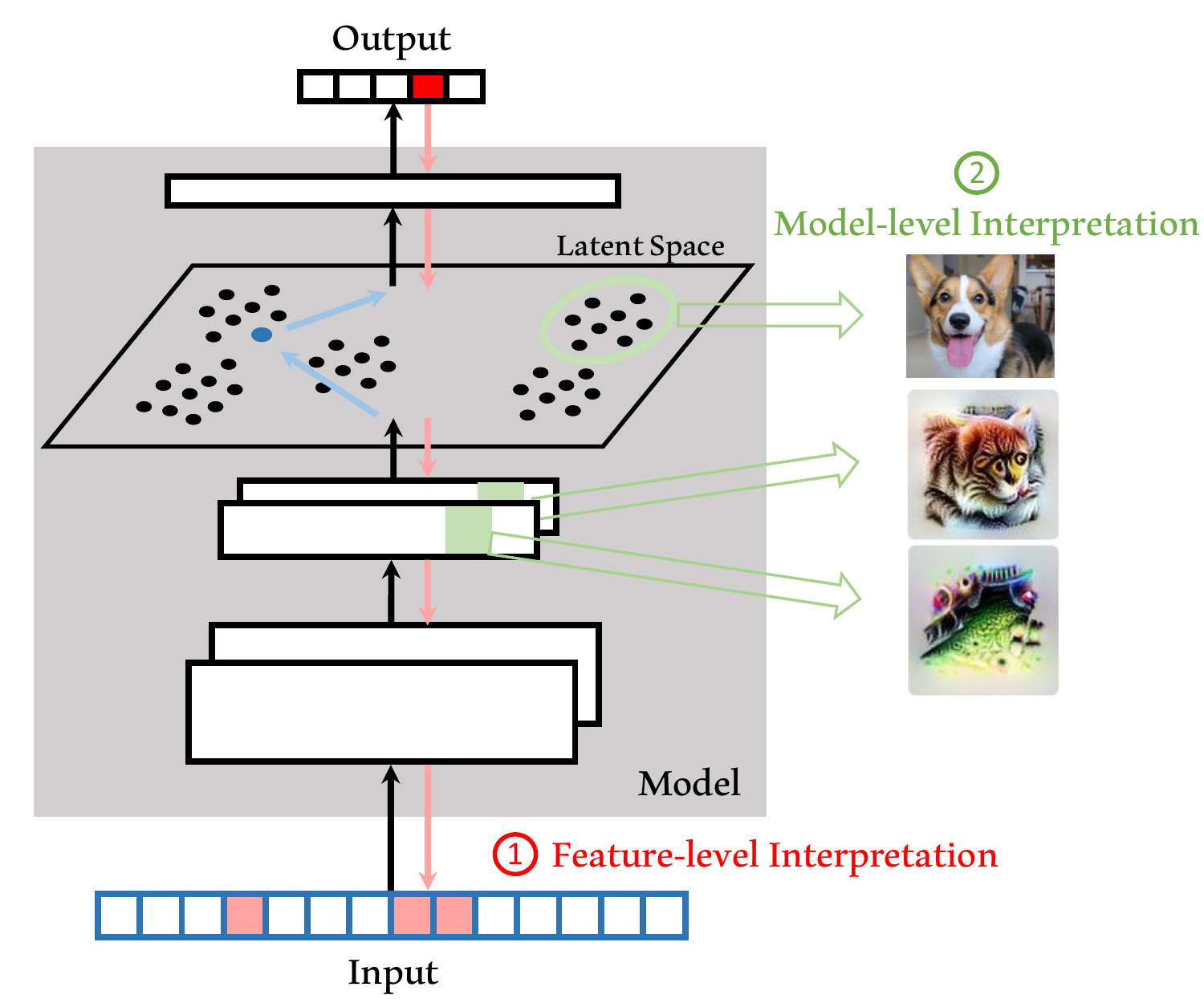}
 \vspace{0pt}
 \caption{ Illustration of Feature-level Interpretation and Model-level Interpretation for a deep model. } \label{fig:intp_category}
\end{figure}

Before connecting the two domains, we first briefly introduce the subjects of interpretation to be covered in this paper. \textit{Interpretability} is defined as ``the ability to explain or to present in understandable terms to a human~\cite{Doshi-Kim17rigorous}". Although a formal definition of interpretation still remains elusive~\cite{Doshi-Kim17rigorous, lombrozo2006structure, keil2006explanation, hempel1948studies}, the overall goal is to obtain and transform information from models or their behaviors into a domain that human can make sense of~\cite{Montavon-etal18methodsSurvey}. For a more structured analysis, we categorize existing work into two categories: feature-level interpretation and model-level interpretation, as shown in Figure~\ref{fig:intp_category}. Feature-level interpretation targets to find the most important features in a data sample to its prediction. Model-level interpretation explores the functionality of model components, and their internal states after being fed with input. This categorization is based on whether the internal working mechanism of models is involved in interpretation.

Following the above categorization, the overall structure of this article is organized as below. To begin with, we briefly introduce different types of adversarial attack and defense strategies in Section~\ref{sec:attack}. Then, we introduce different categories of interpretation approaches, and demonstrate in detail how interpretation correlates to the attack and defense strategies. Specifically, we discuss feature-level interpretation in Section~\ref{sec:intp_feature} and model-level interpretation in Section~\ref{sec:intp_model}. After that, we extend the discussion to additional relations between interpretation and adversarial aspects of model in Section~\ref{sec:additional}. Finally, we discuss some opening challenges for future work in Section~\ref{sec:future}.

\section{Adversarial Machine Learning}\label{sec:attack}
Before understanding how interpretation helps adversarial attack and defense, we first provide an overview of existing attack and defense methodologies.

\subsection{Adversarial Attacks}
In this subsection, we introduce different types of threat models for adversarial attack. The overall threat models may be categorized under different criteria. Based on different application scenarios, conditions, and adversary capabilities, specific attack strategies will be deployed.

\subsubsection{Untargeted vs Targeted Attack} Based on the goal of attackers, the threat models can be classified into targeted and untargeted ones. For \textit{targeted} attack, it attempts to mislead a model's prediction to a specific class given an instance. Let $f$ denote the target model exposed to adversarial attack. A clean data instance is $\textbf{x}_0\in X$, and $X$ is the input space. We consider classification tasks, so $f(\textbf{x}_0) = c, c\in \{1,2,...,C\}$. One way of formulating the task of targeted attack is as below~\cite{Szeg-etal13intriguing}:
\begin{equation}\label{eq:def_targeted}
    \min_{\textbf{x}\in X} \,\, d(\textbf{x}, \textbf{x}_0), \,\,\,\, \text{s.t.} \,\, f(\textbf{x}) = c'
\end{equation}
where $c'\neq c$, and $d(\textbf{x}, \textbf{x}_0)$ measures the distance between the two instances. A typical choice of distance measure is to use $l_p$ norms, where $d(\textbf{x}, \textbf{x}_0) = \|\textbf{x} - \textbf{x}_0\|_p$. The core idea is to add small perturbation to the original instance $\textbf{x}_0$ to make it being classified as $c'$. However, in some cases, it is important to increase the confidence of perturbed samples being misclassified, so the task may also be formulated as:
\begin{equation}\label{eq:def_targeted2}
    \max_{\textbf{x}\in X} \,\, f_{c'}(\textbf{x}), \,\,\,\, \text{s.t.} \,\, d(\textbf{x}, \textbf{x}_0)\le \delta
\end{equation}
where $f_{c'}(\textbf{x})$ denotes the probability or confidence that $\textbf{x}$ is classified as $c'$ by $f$, and $\delta$ is a threshold limiting perturbation magnitude.
For \textit{untargeted} attack, its goal is to prevent a model from assigning a specific label to an instance. The objective of untargeted attack could be formulated in a similar way as targeted attack, where we just need to change the constraint as $f(\textbf{x})\neq c$ in Equation~\ref{eq:def_targeted}, or change the objective as $\min_{\textbf{x}\in X} f_{c}(\textbf{x})$ in Equation~\ref{eq:def_targeted2}. 

In some scenarios, the two types of attack above are also called \textit{false positive} attack and \textit{false negative} attack. The former aims to make models misclassify negative instances as positive, while the latter tries to mislead models to classify positive instances as negative. False positive attack and false negative attack sometimes are also called Type-I attack and Type-II attack.

\subsubsection{One-Shot vs Iterative Attack}
According to practical constraints, adversaries may initiate one-shot or iterative attack to target models. In \textit{one-shot} attack, they have only one chance to generate adversarial samples, while \textit{iterative attack} could take multiple steps to explore better direction. Iterative attack can generate more effective adversarial samples than one-shot attack. However, it also requires more queries to the target model and more computation to initiate each attack, which may limit its application in some computational-intensive tasks.

\subsubsection{Data Dependent vs Universal Attack}
According to information sources, adversarial attacks could be data dependent or independent. In \textit{data dependent} attack, perturbations are customized based on the target instance. For example, in Equation~\ref{eq:def_targeted}, the adversarial sample $\textbf{x}$ is crafted based on the original instance $\textbf{x}_0$. However, it is also possible to generate adversarial samples without referring to the input instance, and it is also named as \textit{universal} attack~\cite{Moosavi-etal17universalPerturb, Metzen-etal17detectPerturbation}. The problem can be abstracted as looking for a perturbation vector $\textbf{v}$ so that
\begin{equation}
    f(\textbf{x} + \textbf{v}) \neq f(\textbf{x}) \,\,\, \text{for ``most"} \,\,\, \textbf{x}\in X .
\end{equation}
We may need a number of training samples to obtain $\textbf{v}$, but it does not rely on any specific input at test time. Adversarial attack can be implemented efficiently once the vector $\textbf{v}$ is solved.

\subsubsection{Perturbation vs Replacement Attack}
Adversarial attacks can also be categorized based on the way of input distortion. In \textit{perturbation} attack, input features are shifted by specific noises so that the input is misclassified by the model. In this case, let $\textbf{x}^*$ denote the final adversarial sample, then it can be obtained via
\begin{equation}
    \textbf{x}^* = \textbf{x}_0 + \Delta \textbf{x},
\end{equation}
and usually $\|\Delta \textbf{x}\|_p$ is small.

In \textit{replacement} attack, certain parts of input are replace by adversarial patterns. Replacement attack is more natural in physical scenarios. For examples, criminals may want to wear specifically designed glasses to prevent them from being recognized by computer vision systems~\footnote{https://www.inovex.de/blog/machine-perception-face-recognition/}. Also, surveillance cameras may fail to detect persons wearing clothes attached with adversarial patches~\cite{Thys-Ranst19surveilanceCamera}. Suppose $\textbf{v}$ denote the adversarial pattern, then replacement attack can be represented by using a mask $\textbf{m}\in \{0,1\}^{|\textbf{x}_0|}$, so that
\begin{equation}
    \textbf{x}^* = \textbf{x}_0\odot (\textbf{1}-\textbf{m}) + \textbf{v}\odot \textbf{m}
\end{equation}
where the symbol $\odot$ denotes element-wise multiplication.

\subsubsection{White-Box vs Black-Box Attack}
In \textit{white-box} attack, it is assumed that attackers know everything about the target model, which may include model architecture, weights, hyper-parameters and even training data. White-box attack helps discovering intrinsic vulnerabilities of the target model. It works in ideal cases representing the worst scenario that defenders have to confront. \textit{Black-box} attack assumes that attackers are only accessible to the model output, just like normal end users. This is a more practical assumption in real-world scenarios. Although a lot of detailed information about models are occluded, black-box attack still poses significant threat to machine learning systems due to the transferability property of adversarial samples discovered in~\cite{Pape-etal16transferability}. In this sense, attacker could build a new model $f'$ to approximate the target model $f$, and adversarial samples created on $f'$ could still be effective to $f$.
\subsection{Defenses against Adversarial Attacks}
In this subsection, we briefly introduce the basic idea of different defense strategies against adversaries.

\subsubsection{Input Denoising}
As adversarial perturbation is a type of human imperceptible noise added to data, then a natural defense solution is to filter it out, or to use additional random transformation to offset adversarial noise. It is worth noting that $f_m$ could be added prior to model input layer~\cite{Xie-etal18randomization, Liao-etal18NIPSChallenge, Xu-etal17feature}, or as an internal component inside the target model~\cite{Xie-etal18denoising}. Formally, for the former case, given an instance $\textbf{x}^*$ which is probably affected by adversaries, we hope to design a mapping $f_m$, so that $f(f_m(\textbf{x}^*))= f(\textbf{x}_0)$. For the latter case, the idea is similar except that $f$ is replace by certain intermediate layer output $h$.

\subsubsection{Model Robustification}
Refining the model to prepare itself against potential threat from adversaries is another widely applied strategy. The refinement of model could be achieved from two directions: changing the training objective, or modifying model structure. Some examples of the former include adversarial training~\cite{Szeg-etal13intriguing, Good-etal15explaining}, and replacing empirical training loss with robust training loss~\cite{Madry-etalICLR18deepResistant}. The intuition behind is to consider in advance the threat of adversarial samples during model training, so that the resultant model gains robustness from training. Examples of model modification include model distillation~\cite{Pape-etal16distillation}, applying layer discretization~\cite{Lu-etal17detectingRejecting}, controlling neuron activations~\cite{Tao-etalNIPS18attackMeetsInterp}. Formally, let $f'$ denote the robust model, the goal is to make $f'(\textbf{x}^*))= f'(\textbf{x}_0)=y$.

\subsubsection{Adversarial Detection}
Different from the previous two strategies where we hope to discover the true label given an instance, adversarial detection tries to identify whether the given instance is polluted by adversarial perturbation. The general idea is to build another predictor $f_d$, so that $f_d(\textbf{x})=1$ if $\textbf{x}$ has been polluted, and otherwise $f_d(\textbf{x})=0$. The establishment process of $f_d$ could follow the normal routine of building a binary classifier~\cite{Gong-etal17notTwins, Meng-Chen17magnet, Grosse-etal17statisticDetection}.

Input denoising and model robustification methods proactively recover the correction prediction from influences of adversarial attack, by fixing the input data and model architectures respectively. Adversarial detection methods reactively decide whether the model should make predictions against the input in order not to be fooled. Implementations of the proactive strategies are usually more challenging than the reactive one.

\section{Feature-Level Interpretation in Adversarial Machine Learning}\label{sec:intp_feature}
Feature-level interpretation is a widely used post-hoc method to identify feature importance with respect to a prediction result. It focuses on the end-to-end relation between input and output, instead of carefully examining the internal states of models. Some examples include measuring the importance of phrases of sentences in text classification~\cite{du2019attribution}, and pixels in image classification~\cite{Zhou-etal16CAM}. In this section, we will discuss how this type of interpretation correlates with the attack and defense of adversaries, although most work on adversarial machine learning does not analyze adversaries from this perspective.

\subsection{Feature-Level Interpretation for Understanding Adversarial Attack}\label{sec:inst_intp_4_adv}
In this part, we will show that many feature-level interpretation techniques are closely coupled with existing adversarial attack methods, thus providing another perspective to understand adversarial attack.

\subsubsection{Gradient-Based Techniques}
Following the notations in previous discussion, we let $f_c(\textbf{x}_0)$ denote the probability that model $f$ classify the input instance $\textbf{x}_0$ as class $c$. One of the intuitive ways to understand why such prediction is derived is to attribute prediction $f_c(\textbf{x}_0)$ to feature dimensions of $\textbf{x}_0$. A fundamental technique to obtain attribution scores is backpropagation. According to~\cite{Simonyan-etal13deepInsideCNNsaliency}, $f_c(\textbf{x}_0)$ can be approximated with a linear function surrounding $\textbf{x}_0$ by computing its first-order Taylor expansion:
\begin{equation}
    f_c(\textbf{x}) \approx f_c(\textbf{x}_0) + \textbf{w}_c^T \cdot (\textbf{x}-\textbf{x}_0)
\end{equation}
where $\textbf{w}_c$ is the gradient of $f_c$ with respect to input at $\textbf{x}_0$, i.e., $\textbf{w}_c = \nabla_{\textbf{x}} f_c(\textbf{x}_0)$. From the interpretation perspective, $\textbf{w}_c$ entries of large magnitude correspond to the features that are important around the current output. 

However, another perspective to comprehend the above equation is that, the interpretation coefficient vector $\textbf{w}_c$ also indicates the most effective direction of locally changing the prediction result by perturbing input away from $\textbf{x}_0$. If we let $ \Delta \textbf{x} = \textbf{x}-\textbf{x}_0 = -\textbf{w}_c$, we are attacking the model $f$ with respect to the input-label pair $(c, \textbf{x}_0)$. Such perturbation method is closely related to the Fast Gradient Sign (FGS) attacking method~\cite{Good-etal15explaining}, where:
\begin{equation}
    \Delta \textbf{x} = \epsilon \cdot sign(\nabla_{\textbf{x}} J(f, \textbf{x}_0, c)),
\end{equation}
except that (1) FGS computes the gradient of a certain cost function $J$ nested outside $f$, and (2) it applies an additional $sign()$ operation on gradient for processing images. However, if we define $J$ with cross entropy loss, and the true label of $\textbf{x}_0$ is $c$, then 
\begin{equation}
    \nabla_{\textbf{x}} J(f, \textbf{x}_0, c) = -\nabla_{\textbf{x}} \log f_c(\textbf{x}_0) = -\frac{1}{f_c(\textbf{x}_0)} \nabla_{\textbf{x}} f_c(\textbf{x}_0),
\end{equation}
which points to the same perturbation direction as reversing the interpretation $\textbf{w}_c$. Both gradient-based interpretation and FGS rely on the assumption that the targeted model can be locally approximated by linear models.

The traditional FGS method is proposed under the setting of untargeted attack, where the goal is to impede input from being correctly classified. For targeted attack, where the goal is to misguide the model prediction towards a specific class, a typical way is Box-constrained L-BFGS (L-BFGS-B) method~\cite{Szeg-etal13intriguing}. Assume $c'$ is the target label, the problem of L-BFGS-B is formulated as:
\begin{equation}
    \argmin_{\textbf{x}\in X}\,\,\, \alpha \cdot d(\textbf{x}, \textbf{x}_0) + J(f, \textbf{x}, c')
\end{equation}
where $d$ is considered to control perturbation degree, and $X$ is the input domain (e.g., $[0,255]$ for each channel of image input). The goal of attack is to make $f(\textbf{x})=c'$, while maintaining $d(\textbf{x}, \textbf{x}_0)$ to be small. Suppose we apply gradient descent to solve the problem, and $\textbf{x}_0$ is the starting point. Similar to the previous discussion, if we define $J$ as the cross entropy loss, then
\begin{equation}
    -\nabla_{\textbf{x}} J(f, \textbf{x}_0, c')) = \nabla_{\textbf{x}} \log f_{c'}(\textbf{x}_0) \propto \textbf{w}_{c'} \,.
\end{equation}
On one hand, $\textbf{w}_{c'}$ locally and linearly interprets $f_{c'}(\textbf{x}_0)$, and it also serves the most effective direction to make $\textbf{x}_0$ towards being classified as $c'$. 

According to the taxonomy of adversarial attacks, the two scenarios discussed above can also be categorized into: (1) one-shot attack, since we only perform interpretation once, (2) data-dependent attack, since the perturbation direction is related with $\textbf{x}_0$, (3) white-box attack, since model gradients are available. Other types of attack could be crafted if different interpretation strategies are applied, which will be discussed in later sections.

\textbf{Improved Gradient-Based Techniques.} The interpretation methods based on raw gradients, as discussed above, are usually unstable and noisy~\cite{Rudin19pleaseStop, Murdoch-etal19defMethApp}. The possible reasons include: (1) the target model itself is not stable in terms of function surface or model establishment; (2) gradients only consider the local output-input relation so that its scope is too limited; (3) the prediction mechanism is too complex to be approximated by a linear substitute. Some approaches for improving interpretation (i.e., potential adversarial attack) are as below.
\begin{itemize}[leftmargin=*]
    \item \textbf{Region-Based Exploration:} To reduce random noises in interpretation, SmoothGrad is proposed in~\cite{Smilkov-etal18smoothgrad}, where the final interpretation $\textbf{w}_c$, as a sensitivity map, is obtained by averaging a number of sensitivity maps of instances sampled around the target instance $\textbf{x}_0$, i.e., $\textbf{w}_c = \sum_{\textbf{x}' \in \mathcal{N}(\textbf{x}_0)} \, \frac{1}{|\mathcal{N}(\textbf{x}_0)|} \nabla f_c(\textbf{x}')$. The averaged sensitivity map will be visually sharpened. A straightforward way to extend it for adversarial attack is to perturb input by reversing the averaged map. Furthermore,~\cite{Tramer-etal18ensembleAdvTraining} designed a different strategy by adding a step of random perturbation before gradient computation in attack, to jump out of the non-smooth vicinity of the initial instance. Spatial averaging is a common technique to stabilize output. For example, ~\cite{Cao-Gong17regionClassification} applied it as a defense method to derive more stable model predictions.
    \item \textbf{Path-Based Integration:} To improve interpretation, ~\cite{Sundararajan-etal16integratedGradient} proposes Integrated Gradient (InteGrad). After setting a baseline point $\textbf{x}^b$, e.g., a black image in object recognition tasks, the interpretation is defined as:
\begin{equation}
    \textbf{w}_c = \frac{(\textbf{x}_0 - \textbf{x}^b)}{D} \circ \sum_{d=1}^D \, [\nabla f_c](\textbf{x}^b + \frac{d}{D}(\textbf{x}_0 - \textbf{x}^b)) ,
\end{equation}
which is the weighted sum of gradients along the straight-line path from $\textbf{x}_0$ to the baseline point $\textbf{x}^b$. 
A similar strategy in adversarial attack is iterative attack~\cite{Kurakin-etal17atScale}, where the sample is iteratively perturbed as:
\begin{equation}
    \textbf{x}'_0 = \textbf{x}_0, \,\,\, \textbf{x}'_{d+1} = Clip_{\textbf{x}_0, \epsilon}\{ \textbf{x}'_d + \alpha \nabla_{\textbf{x}} J(f, \textbf{x}'_d, c) \} ,
\end{equation}
which gradually explore the perturbation along a path directed by a series of gradients. $Clip_{\textbf{x}_0, \epsilon}(\textbf{x})$ denotes element-wise clipping $\textbf{x}$ so that $d(\textbf{x} - \textbf{x}_0) \le \epsilon$.
\end{itemize}

Interestingly, although in many cases gradient-based interpretation is intuitive as visualization to show that the model is functioning well, it may be an illusion since we can easily transform interpretation into adversarial perturbation.

\subsubsection{Distillation-Based Techniques}
The interpretation techniques discussed so far require gradient information $\nabla_{\textbf{x}} f$ from models. Meanwhile, it is possible to extract interpretation without querying a model $f$ more than $f(\textbf{x})$. This type of interpretation methods, here named as distillation-based methods, can be also used for adversarial attack. Since no internal knowledge is required from the target model, they are usually used for black-box attack.

The main idea of applying distillation for interpretation is to use an interpretable model $g$ (e.g., a decision tree) to mimic the behavior of the target deep model $f$~\cite{Che-etal15distilling, gao2017interpretable}. Once we obtain $g$, existing white-box attack methods could be applied to craft adversarial samples~\cite{Liu-etal18adversarial}. In addition, given an instance $\textbf{x}_0$, to guarantee that $g$ more accurately mimics the nuanced behaviors of $f$, we could further require that $g$ locally approximates $f$ around the instance. The objective is thus as below:
\begin{equation}
    \min_g \mathcal{L}(f, g, \textbf{x}_0) + \alpha \cdot C(g),
\end{equation}
where $\mathcal{L}$ denotes the approximation error around $\textbf{x}_0$. For examples, in LIME~\cite{Ribe-etal16why}:
\begin{equation}
    \mathcal{L}(f, g, \textbf{x}_0) = \sum_{\textbf{x}' \in \mathcal{N}(\textbf{x}_0)} \exp(-d(\textbf{x}_0, \textbf{x}')) \|f(\textbf{x}') - g(\textbf{x}')\|^2 ,
\end{equation}
and $\mathcal{N}(\textbf{x}_0)$ denotes the local region around $\textbf{x}_0$. In addition, LEMNA~\cite{Guo-etal18lemna} adopts mixture regression models for $g$ and fused lasso as regularization $C(g)$.
After obtaining $g$, we can craft adversarial samples targeting $g$ using attack methods by removing or reversing the interpretation result.
According to the property of transferability~\cite{Pape-etal16transferability}, an adversarial sample that successfully fools $g$ is also likely to fool $f$. The advantages are two-fold. First, the process is model agnostic and does not assume availability to gradients. It could be used for black-box attack or attacking certain types of models (such as tree-based models) that do not use gradient backpropagation in training. Second, one-shot attacks on $g$ could be more effective thanks to the smoothness term $C(g)$ as well as extending the consideration to include the neighborhood of $\textbf{x}_0$~\cite{Bigg-etal13evasion}. Thus, it has the potential to cause defense methods that are based on obfuscated gradients~\cite{Athalye-etal18obfuscatedGradientFalse} to be less robust. The disadvantage is that crafting each adversarial sample requires high computation cost.

In certain scenarios, it is beneficial to make adversarial patterns understandable to humans as real-world simulation when identifying model vulnerability. For examples, in autonomous driving, we need to consider physically-possible patterns that could cause misjudgement of autonomous vehicles~\cite{Boloor-etal19physicalAutoDrive}. One possible approach is to constrain adversarial instances to fall into the data distribution. For example, \cite{Dhurandhar-etalNIPS18constrastive} achieves this through an additional regularization term $\|  \textbf{x}_0 + \Delta \textbf{x} - AE(\textbf{x}_0 + \Delta \textbf{x}) \|$, where $AE(\cdot)$ denotes an autoencoder. Another strategy is to predefine a dictionary, and then makes the adversarial perturbation to match one of the dictionary tokens~\cite{Boloor-etal19physicalAutoDrive}, or a weighted combination of the tokens~\cite{Sato-etal18interpretableAdvText}.

\begin{figure}[t]
\centering
 \vspace{0pt}
 \includegraphics[width=0.40\textwidth]{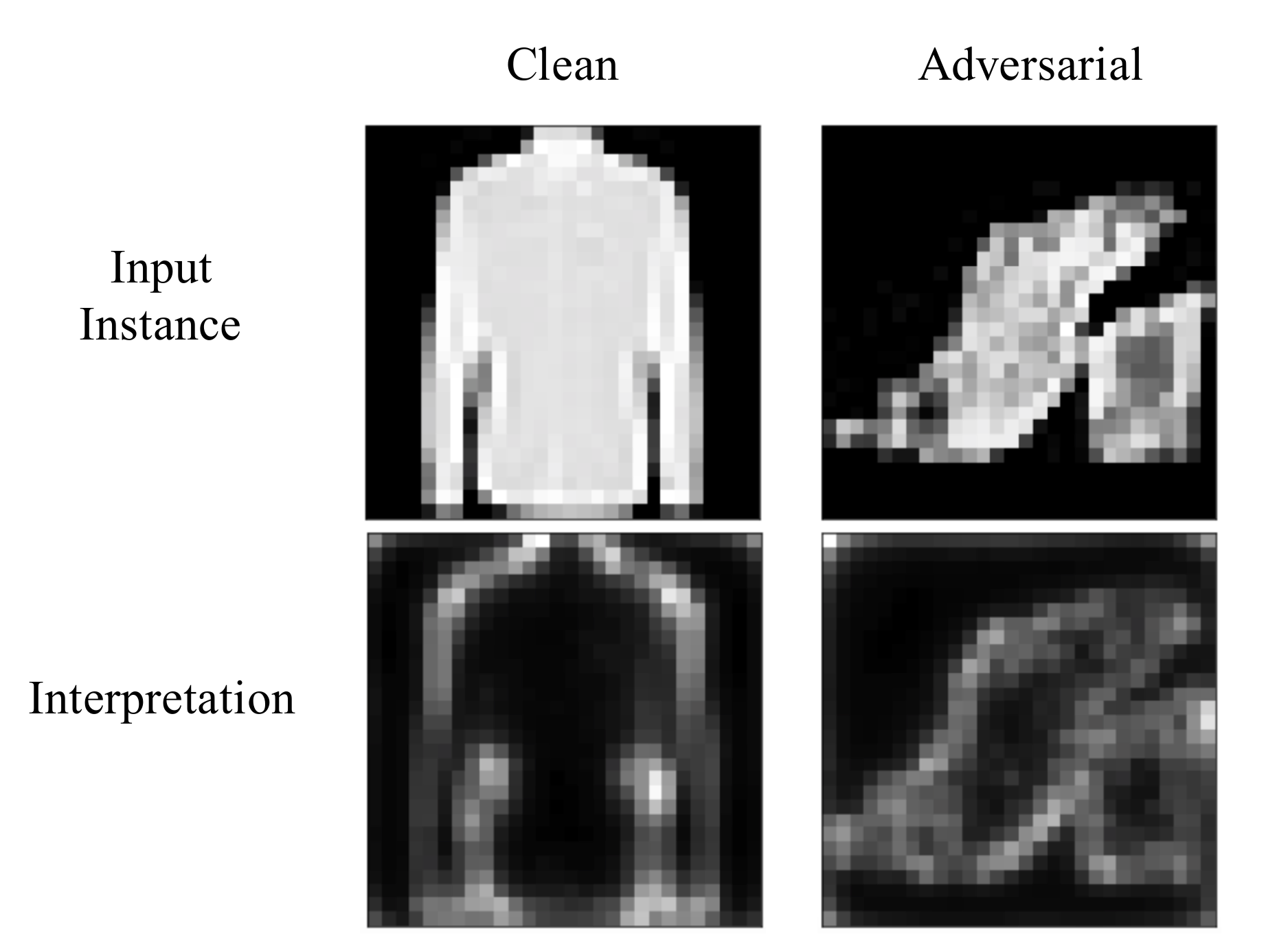}
 \vspace{0pt}
 \caption{The interpretation of an adversarial sample may differ from the one of a clean sample. Top-left: a normal example from the shirt class of Fashion-MNIST dataset. Bottom-left: explanation map for the classification. Top-right: an adversarial example, originally from the sandal class, that is misclassified as a shirt. Bottom-right: explanation map for the misclassification.} \label{fig:adv_intp_compare}
 \vspace{0pt}
\end{figure}

\subsection{Feature-level Interpretation against Adversaries}
Feature-level interpretation could be used for defense against adversaries through adversarial training and detecting model vulnerability.

\subsubsection{Model Robustification with Feature-level \\Interpretation}
Feature-level interpretation could help adversarial training to improve model robustness. Adversarial training~\cite{Good-etal15explaining, Kurakin-etal17atScale} is one of the most applied proactive countermeasures to improve the robustness of the model. Its core idea is to first generate adversarial samples to unveil the weakness of the model, and then inject the adversarial samples into training set for data augmentation. The overall loss function can be formulated as:
\begin{equation}
    \min_f\, \mathbb{E}_{(\textbf{x}, y)\in \mathcal{D}}\, [\alpha J(f(\textbf{x}), y) + (1-\alpha)J(f(\textbf{x}^*), y)].
\end{equation}
In the scenario of adversarial training, feature-level interpretation helps in preparing adversarial samples $\textbf{x}^*$, which may refer to any method discussed in Section~\ref{sec:inst_intp_4_adv}. Although such an attack-and-then-debugging strategy has been successfully applied in many traditional cybersecurity scenarios, one key drawback is that it tends to overfit to the specific approach that is used to generate $\textbf{x}^*$. It is untenable and ineffective~\cite{He-etal17ensembleNotStrong} to exhaust a number of possible attacking methods for data preparation. Meanwhile, it is argued that naive adversarial training may actually perform gradient masking instead of moving the decision boundary~\cite{Athalye-etal18obfuscatedGradientFalse, 0NIPS2017}. 

To train more robust models, some optimization based methods have been proposed.~\cite{Madry-etalICLR18deepResistant} argued that traditional Empirical Risk Minimization (ERM) fails to yield models that are robust to adversarial instances, and proposed a min-max formulation to train robust models: 
\begin{equation}
    \min_f\, \mathbb{E}_{(\textbf{x}, y)\in \mathcal{D}}\, [\max_{\delta\in \Delta X} J(\textbf{x}+\delta, y)] ,
\end{equation}
where $\Delta X$ denotes the set of allowed perturbations.
It formally defines adversarially robust classification as a learning problem to reduce adversarial expected risk. This min-max formulation provides another perspective on adversarial training, where the inner task aims to find adversarial samples, and the outer task retrains model parameters.~\cite{Tramer-etal18ensembleAdvTraining} further improves its defense performance by crafting adversarial samples from multiple sources to augment training data. This strategy is also implicitly supported in~\cite{Schmidt-etal18moredata} which shows training robust models requires much greater data complexity. \cite{Zhang-etal19theoretically} further identifies a trade-off between robust classification error~\cite{Schmidt-etal18moredata, Cullina-etal18paclearning} and natural classification error, which provides a solution to reduce the negative effect on model accuracy after adversarial training.

Besides adversarial training, feature-level interpretation can also provide motivation to robust learning. For example, empirical interpretation results pointed out, that an intriguing property of CNN is its bias towards texture instead of shape in making predictions~\cite{baker2018deep}. To tackle this problem, \cite{Shi-etal20infodrop} proposes InfoDrop, a plug-in filtering method to remove texture-intensive information during forward propagation of CNN. Feature map regions with low self-information, i.e., regions being observed based on their contents contain less ``surprise", tend to be filtered out. In this way, the model will pay more attention to regions such as edges and corners, and be more robust under various scenarios including adversaries.


\subsubsection{Adversarial Detection with Feature-level \\Interpretation}
In the scenario where a model is subject to adversarial attack, interpretation may serve as a new type of information for directly detecting adversarial patterns. The motivation is illustrated in Figure~\ref{fig:adv_intp_compare}. For the adversarial image which originally shows a shoe, although the model classifies it as shirt, its interpretation result does not resemble the one obtained from the clean image of a shirt. A straightforward way to distinguish interpretations is to train another classifier $f_d$ as the detector trained with interpretations of both clean and adversarial instances, paired with labels indicating whether the sample is clean~\cite{Fidel-etal19detectSHAP, Fong-Vedaldi17perturbation, Yang-etal19detectAttribution, Zhang-etal18detectSaliency}. Specifically, \cite{Zhang-etal18detectSaliency} directly uses gradient-based saliency map as interpretation, \cite{Yang-etal19detectAttribution} adopts the distribution of Leave-One-Out (LOO) attribution scores, while \cite{Fong-Vedaldi17perturbation} proposes a new interpretation method based on masks highlighting important regions. \cite{Wang-etal20interpSafety} proposes an ensemble framework called X-Ensemble for detecting adversarial samples. X-Ensemble consists of multiple sub-detectors, each of which is a convolutional neural network to classify whether an instance is adversarial or benign. The input to each sub-detector is the interpretation on the instance's prediction. More than one interpretation methods are deployed so there are multiple sub-detectors. A random forest model is then used to combine sub-detectors into a powerful ensemble detector.

In more scenarios, interpretation serves as a diagnosis tool to qualitatively identify model vulnerability. First, we could use interpretation to identify whether inputs are affected by adversarial attack. For example, if interpretation result shows that unreasonable evidences have been used for prediction~\cite{Dong-etal17improveInterpretWithAdversarial}, then it is possible that there exists suspicious but imperceptible input pattern. Second, interpretation may reflect whether a model is susceptible to adversarial attack. Even given a clean input instance, if interpretation of model prediction does not make much sense to human, then the model is under the risk of being attacked. For examples, in a social spammer detection system, if the model regards certain features as important but they are not strongly correlated with maliciousness, then attackers could easily manipulate these features without much cost to fool the system~\cite{Liu-etal18adversarial}. Also, in image classification, CNN models have been demonstrated to focus on local textures instead of object shapes, which could be easily utilized by attackers~\cite{baker2018deep}. An interesting phenomenon in image classification is that, after refining a model with adversarial training, feature-level interpretation results indicate that the refined model will be less biases towards texture features~\cite{Zhang-Zhu19interpretAdvTrained}.

Nevertheless, there are several challenges that impede the intuitions above from being formulated to formal defense approaches. First, interpretation itself is also fragile in neural networks. Attackers could control prediction and interpretation simultaneously via indistinguishable perturbation~\cite{Ghorbani-etal19fragile, Subramanya-etal19foolingIntp}. Second, it is difficult to quantify the robustness of model through interpretation~\cite{Rudin19pleaseStop}. Manual inspection of interpretation helps discover defects in model, but visually acceptable interpretation does not guarantee model robustness. That is, defects in feature-level interpretation indicate the presence but not the absence of vulnerability.




\section{Model-level Interpretation in Adversarial Machine Learning}\label{sec:intp_model}
In this review, model-level interpretation is defined with two aspects. First, model-level interpretation aims to figure out what has been learned by intermediate components in a trained model~\cite{ZhangQ-etal18cnnTrees, Simonyan-etal13deepInsideCNNsaliency}, or what is the meaning of different locations in latent space~\cite{Kim-etal18concepts, Zhou-etal18visualBasis, liu2018interpretation}. Second, given an input instance, model-level interpretation unveils how the input is encoded by those components as latent representation~\cite{Kim-etal18concepts, Zhou-etal18visualBasis, Liao-etal18NIPSChallenge, Xie-etal18denoising}. In our discussion, the former does not rely on input instances, while the later is the opposite. Therefore, we name the two aspects as \textit{Static Model Interpretation} and \textit{Representation Interpretation} respectively to further distinguish them. Representation interpretation could rely on static model interpretation.

\subsection{Static Model Interpretation for Understanding Adversarial Attack}
For deep models, one of the most widely explored strategies is to explore the visual or semantic meaning of each neuron. A popular strategy for solve this problem is to recover the patterns that activate the neuron of interests at a specific layer~\cite{Erhan-etal09visualizingHigherLayers, Simonyan-etal13deepInsideCNNsaliency}. Following the previous notations, let $h(\textbf{x})$ denote the activation of neuron $h$ given input, the perceived pattern of the neuron can be visualized via solving the problem below:
\begin{equation}\label{eq:model_intp}
    \argmax_{\textbf{x}'} h(\textbf{x}') - \alpha \cdot C(\textbf{x}'),
\end{equation}
where $C(\cdot)$ such as $\|\cdot\|_1$ or $\|\cdot\|_2$ acts as regularization. Conceptually, the result contains patterns that neuron $h$ is sensitive to. If we choose $h$ to be $f_c$, then the resultant $\textbf{x}'$ illustrates class appearances learned by the target model. Another discussion about different choices of $h$, such as neurons, channels, layers, logits and class probabilities, is provided in~\cite{olah2017feature}.
Similarly, we could also formulate another minimization problem
\begin{equation}\label{eq:model_intp_reverse}
    \argmin_{\textbf{x}'} h(\textbf{x}') + \alpha \cdot C(\textbf{x}'),
\end{equation}
to produce patterns that prohibit activation of certain model components or prediction towards certain classes.

\begin{figure}[t]
\centering
 \vspace{0pt}
 \includegraphics[width=0.48\textwidth]{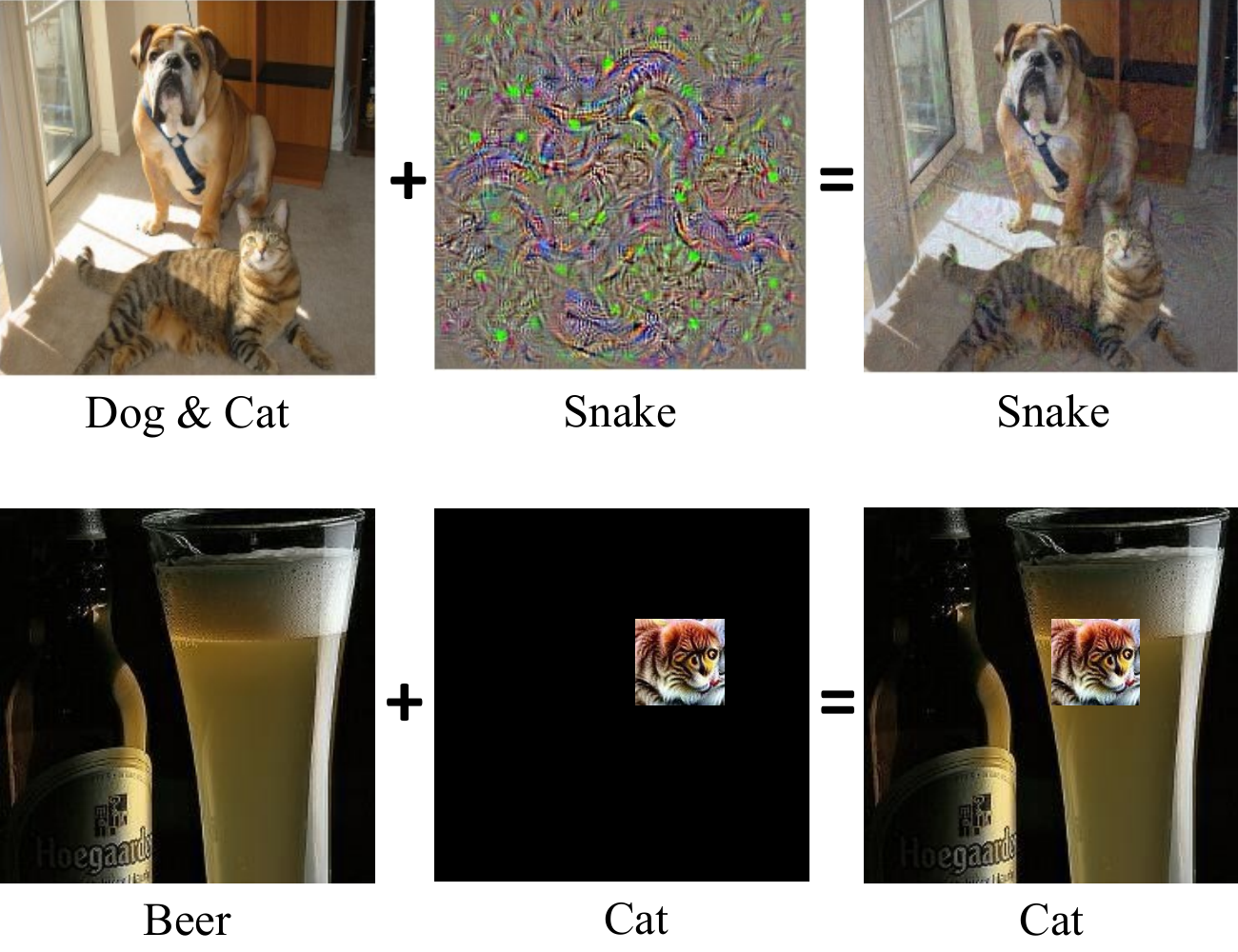}
 \vspace{0pt}
 \caption{ Example of adversarial attack after applying model-level interpretation. Upper: Targeted universal perturbation. Lower: Universal replacement attack.} \label{fig:model_intp_adv}
\end{figure}

The interpretation result $\textbf{x}'$ is highly related with several types of adversarial attack, with some examples shown in Figure~\ref{fig:model_intp_adv}:
\begin{itemize}[leftmargin=*]
    \item \textbf{Targeted-Universal-Perturbation Attack:} If we set $h$ to be class relevant mapping such as $f_c$, then $\textbf{x}'$ can be directly added to target input instance as targeted perturbation attack. That is, given a clean input $\textbf{x}_0$, the adversarial sample $\textbf{x}^*$ is crafted simply as $\textbf{x}^*=\textbf{x}_0 + \lambda\cdot \textbf{x}'$ to make $f(\textbf{x}^*)=c$. It belongs to universal attack, because the interpretation process in Eq.\ref{eq:model_intp} does not utilize any information of the clean input. 
    \item \textbf{Untargeted-Universal-Perturbation Attack:} If we set $h$ to be the aggregation of a number of middle-level layer mappings, such as $h(\textbf{x}') = \sum_l \log(h^l(\textbf{x}'))$ where $h^l$ denotes the feature map tensor at layer $l$, the resultant $\textbf{x}'$ is expected to produce spurious activation to confuse the prediction of CNN models given any input, which implies $f(\textbf{x}_0+ \lambda\cdot \textbf{x}')\neq f(\textbf{x}_0)$ with high probability~\cite{Mopuri-etal17universalDataIndependent}.
    \item \textbf{Universal-Replacement Attack:} Adversarial patches, which completely replace part of input, represent a visually different attack from perturbation attack. Based on Eq.\ref{eq:model_intp}, more parameters such as masks, shape, location and rotation could be considered in the optimization to control $\textbf{x}'$~\cite{Brown-etal18advPatch}. The patch is obtained as $\textbf{x}'\odot \textbf{m}$, and the adversarial sample $\textbf{x}^* = \textbf{x}_0\odot (\textbf{1}-\textbf{m}) + \textbf{x}'\odot \textbf{m}$, where $\textbf{m}$ is a binary mask that defines patch shape. Besides, based on Eq.\ref{eq:model_intp_reverse}, after defining $h$ as the objectness score function in person detectors~\cite{Thys-Ranst19surveilanceCamera} or as the logit corresponding to human class~\cite{Sharif-etal16glasses}, it produces real-world patches attachable to human bodies to avoid them being detected by surveillance camera.
\end{itemize}

\subsection{Representation Interpretation for Initiating Adversarial Attack}
Representation learning plays a crucial role in recent advances of machine learning, with applications in vision~\cite{bengio2013representation}, natural language processing~\cite{Young-etal18survey} and network analysis~\cite{Hamilton-etal18survey}. However, the opacity of representation space also becomes the bottleneck for understanding complex models. A commonly used strategy toward understanding representation is to define a set of explainable basis, and then decompose representation points along the basis. Formally, let $\textbf{z}_i\in \mathbb{R}^D$ denote a representation vector, and $\{\textbf{b}_k\in \mathbb{R}^D\}^K_{k=1}$ denote the basis set, where $D$ denotes the representation dimension and $K$ is the number of base vectors. Then, through decomposition
\begin{equation}
    \textbf{z}_i = \sum^K_{k=1} p_{i, k} \cdot \textbf{b}_k ,
\end{equation}
we can explain the meaning of $\textbf{z}_i$ through referencing base vectors whose semantics are known, where $p_{i, k}$ measures the affiliation degree between instance $\textbf{z}_i$ and $\textbf{b}_k$. The work of providing representation interpretation following this scheme can be divided into several groups:
\begin{itemize}[leftmargin=*]
    \item \textbf{Dimension-wise Interpretation}: A straightforward way to achieve interpretability is to require each dimension to have a concrete meaning~\cite{Higgins-etal17betaVAE, Panigrahi-etal19word2sense}, so that the basis can be seen as non-overlapping one-hot vectors. A natural extension to this would be to allow several dimensions (i.e., a segment) to jointly encode one meaning~\cite{liu2019single, Ma-etal19disentangled}.
    \item \textbf{Concept-wise Interpretation}: A set of high-level and intuitive concepts could first be defined, so that each $\textbf{b}_k$ encodes one concept. Some examples include visual concepts~\cite{Zhou-etal18visualBasis, Kim-etal18concepts, Ghorbani-etal19automaticConcept}, antonym words~\cite{Mathew-etal19polarInterpWordEmbd}, and network communities~\cite{liu2018interpretation}.
    \item \textbf{Example-wise Interpretation}: Each base vector can be designed to match one data instance~\cite{Kim-etal14thebayesian, Koh-Liang17influencefunction, Velickovic-etal18graphAttention} or part of the instance~\cite{Chen-etal19thisLooksthat}. Those instances are also called prototypes. For examples, a prototype could be an image region~\cite{Chen-etal19thisLooksthat} or a node in networks~\cite{Velickovic-etal18graphAttention}.
\end{itemize}

The extra knowledge obtained from representation interpretation could be used to guide the direction of adversarial perturbation. However, the motivation of this type of work usually is to initiate more meaningful adversaries and then use adversarial training to improve model generalization, but not for the pure purpose of undermining model performance.  For examples, in text mining, \cite{Sato-etal18interpretableAdvText} restricts perturbation direction of each word embedding to be a linear combination of vocabulary word embeddings, which improves model performance in text classification with adversarial training. In network embedding, \cite{Dai-etal19advTrainingGraph} restricts perturbation of a node's embedding towards the embeddings of the node's neighbors in the network, which benefits node classification and link prediction.

\subsection{Model-level Interpretation against \\Adversaries}
Model-level interpretation develops an internal understanding of a model, including its weakness. Defenders could either choose to improve model robustness or develop a detector using internal data representation.

\subsubsection{Model Robustification with Model-level \\Interpretation}
Some high-level features learned by deep models are not robust, which is insufficient to train robust models. A novel algorithm is proposed in~\cite{Ilyas-etal19notBugs} to build datasets of robust features. Let $h:X\rightarrow \mathbb{R}$ denote a transformation function that maps input to a representation neuron. Each instance in the robust dataset $\mathcal{D}_r$ is constructed from the original dataset $\mathcal{D}$ through solving a optimization problem:
\begin{equation}
    \mathbb{E}_{(\textbf{x},y)\in \mathcal{D}_r}[h(\textbf{x})\cdot y]=
    \begin{cases}
      \mathbb{E}_{(\textbf{x},y)\in \mathcal{D}}[h(\textbf{x})\cdot y], & \text{if}\ h\in\mathcal{H}_r \\
      0, & \text{otherwise}
    \end{cases}
\end{equation}
where $\mathcal{H}_r$ denotes the set of features utilized by robust models. In this way, input information that corresponds to non-robust representations are suppressed.

Despite not being directly incorporated in the procedure of model training, inspection of model-level interpretation, especially latent representation, has motivated several defense approaches. Through visualizing feature maps of latent representation layers, the noise led by adversarial perturbation can be easily observed~\cite{Xie-etal18denoising, Liao-etal18NIPSChallenge, Yang-etal19detectAttribution}. With this observation, ~\cite{Xie-etal18denoising} proposes adding denoising blocks between intermediate layers of deep models, where the core function of the denoising blocks are chosen as low-pass filters. \cite{Liao-etal18NIPSChallenge} observed that adversarial perturbation is magnified through feedforward propagation in deep models, and proposed a U-net model structure as denoiser. Furthermore, through neuron pattern visualization, \cite{Wang-etal20highFreq} found that convolutional kernels of CNNs after adversarial training tend to show a more smooth pattern. Based on this observation, they propose to average each kernel weight with its neighbors in a CNN model, in order to to improve the adversarial robustness.

\subsubsection{Adversarial Detection with Model-level \\Interpretation}
Instead of training another large model as detector using raw data, we can also leverage model-level interpretation to detect adversarial instances more efficiently. By regarding neurons as high-level features, readily available interpretation methods such as SHAP~\cite{Lundberg-etal17SHAP} could be applied for feature engineering to build adversarial detector~\cite{Fidel-etal19detectSHAP}. After inspecting the role of neurons in prediction, a number of critical neurons could be selected. A steered model could be obtained by strengthening those critical neurons, while adversarial instances are detected if they are predicted very differently by the original model and steered model~\cite{Tao-etalNIPS18attackMeetsInterp}. Nevertheless, the majority of work on adversarial detection utilizes latent representation of instances without inspecting their meanings, such as directly applying statistical methods on representations to build detectors~\cite{Metzen-etal17detectPerturbation, Li-etal17detectConvFilterStat, Feinman-etal17detectArtifacts} or conducting additional coding steps on activations of neurons~\cite{Lu-etal17detectingRejecting}.

\section{Additional Relations Between Adversary and Interpretation}\label{sec:additional}
In previous context, we have discussed how interpretation could be leveraged in adversarial attack and defense. In this section, we complement this viewpoint by analyzing the role of adversarial aspect of models in defining and evaluating interpretation. In addition, we specify the distinction between the two domains.

\subsection{Improving Interpretation via Building\\ Robust Models}
In previous content, we have discussed the role of interpretation in studying model robustness. From another perspective, improving model robustness also influences interpretation of models. First, the representations learned by robust models tend to align better with salient data characteristics and human perception~\cite{Tsipras-etal19odds}. Therefore, adversarially robust image classifiers are also useful in more sophisticated tasks such as generation, super-resolution and translation~\cite{Santurkar-etal19synthesisSingleClf}, even without relying on GAN frameworks. Also, when attacking a robust classifier, resultant adversarial samples tend to be recognized similarly by the classifier and human~\cite{Tsipras-etal19odds}. In addition, retraining with adversarial samples~\cite{Zhang-Zhu19interpretAdvTrained}, or regularizing gradients to improve model robustness~\cite{Ross-Velez18regularizeGradient}, has been discovered to reduce noises from gradient-based sensitivity maps, and encourage CNN models to focus more on object shapes in making predictions. To take a step further, \cite{Zhu-Li20purification} presents the principle of ``feature purification". The work discovers that dense mixtures of patterns exist in the weights of models trained with clean data using normal gradient descent. The dense pattern mixtures still generalize well when being used to predict normal data, but they are extremely sensitive to small perturbation in input. It then theoretically proves under two-layer neural networks that, through adversarial training, dense pattern mixtures could be removed, as visualized through neuron interpretation.

\subsection{Defining Interpretation With Adversaries}
Some definitions of interpretation are inspired by adversarial perturbation. For feature-level interpretation, to understand the importance of a certain feature $x$, we try to answer a hypothetical question that ``What would happen to the prediction Y, if $x$ is removed or distorted?". This is closely related to \textit{causal inference}~\cite{guo2020survey, moraffah2020causal}, and samples crafted in this way are also called \textit{counterfactual explanations}~\cite{wachter2017counterfactual}. For example, to understand how different words in sentences contribute to downstream NLP tasks, we can erase the target words from input, so that the variation in output indicates whether the erased information is important for prediction~\cite{Li-Jurafsky17erasure}. In image processing, salient regions could be defined as the input parts that most affect the output value when perturbed~\cite{Fong-Vedaldi17perturbation}. Considering that using traditional iterative algorithms to generate masks is time-consuming, Goyal et al.~\cite{Dabkowski-Gal17realtimeSaliency} develops trainable masking models that generate masks in real time. In order to make the masks sharp and precise, the U-Net architecture~\cite{ronneberger2015u} is applied for building the trainable model. Besides, objective function above can also be reformulated with the information bottleneck~\cite{Schulz-etal19restrictingFlow}.

Besides defining feature-level interpretation, the similar strategy can be used to define static model interpretation. Essentially we need to answer the question that ``How the model output will change if we change the component in the model?". The general idea is to treat the structure of a deep model as a causal model~\cite{narendra2018explaining}, or extract human understandable concepts to build a causal model~\cite{Harraton-etal18causalExplain}, and then estimate the causal effect of each model component via causal reasoning. The importance of a component is measured by computing model output changes after the component is removed.

As a natural extension from the discussion above, adversarial perturbation can also be used to evaluate the interpretation result. For examples, after obtaining the important features, and understanding whether they are positively or negatively related to the output, we could remove or distort these features to observe the target model's performance change~\cite{Liu-etal18adversarial, Guo-etal18lemna}. If the target model's performance significantly drops, then we are likely to have extracted correct interpretation.
However, it is worth noting that the evaluation will not be fair if the metric and interpretation methods do not match~\cite{liu2020interpretations}.

\subsection{Uniqueness of Model Explainability from Adversaries}
Despite the common techniques applied for acquiring interpretation and exploring adversary characteristics, some aspects of the two directions put radically different requirements. For examples, some applications require interpretation to be easily understood by human especially by AI novices, such as providing more user-friendly interfaces to visualize and present interpretation~\cite{lim2009and, narayanan2018humans, yang2019xfake}, while adversarial attack requires perturbation to be imperceptible to human. Some work tries to adapt interpretation to fit human cognition habits, such as providing example-based interpretation~\cite{bichindaritz2006case}, criticism mechanism~\cite{Kim-etal16examples} and counter-factual explanation~\cite{Goyal-etal19counterfactual}. Furthermore, generative models could be applied to create content from interpretation~\cite{zhao2019personalized}, where interpretation is post-processed into more understandable content such as dialogue texts. The emphasis of understandability in interpretability is exactly opposite to one of the objectives in adversarial attack, which focuses on crafting perturbation that is too subtle to be perceived by human.

\section{Challenges and Future Work}\label{sec:future}
We briefly introduce the challenges encountered in leveraging interpretation to analyze adversarial robustness of models. Finally, we discuss the future research directions.

\subsection{Model Development with Better \\Explainability}
Although interpretation could provide important directions against adversaries, interpretation techniques with better stability and faithfulness are needed before it could really be widely used as a reliable tool. As one of the challenges, it has been shown that many existing interpretation methods are vulnerable to adversarial attacks~\cite{Ghorbani-etal19fragile, Heo-etal19foolInterp, Subramanya-etal19foolingIntp, slack2020fooling}. A stable interpretation method, given an input instance and a target model, should be able to produce relatively consistent result under the situation that input may be subject to certain noises. As preliminary work, \cite{Dombrowski-etal19manipulate} analyzed the phenomenon from a geometric perspective of decision boundary and proposed a smoothed activation function to replace Relu. \cite{Levine-etal19certifyRobustInterp} proposed a sparsified variant of SmoothGrad~\cite{Smilkov-etal18smoothgrad} in producing saliency maps that is certifiably robust to adversarial attacks.

Besides post-hoc interpretation, another challenge we are facing is how to develop intrinsically interpretable models~\cite{Rudin19pleaseStop}. With intrinsic interpretability, it could be easier for model developers to correct undesirable properties of models. One of the challenges is that requiring interpretability may negatively affect model performance. To tackle the problem, some preliminary work start to explore applying graph-based models, such as proposing relational inductive biases to facilitate learning about entities and their relations~\cite{Battaglia-etal18google}, towards a foundation of interpretable and flexible scheme of reasoning. Novel neural architectures have also been proposed such as capsule networks~\cite{Sabour-Hinton17routing} and causal models~\cite{Peters-etal17elements}. 

\subsection{Adversarial Attack in Real-World \\Scenarios}
The most common scenario in existing work considers adversarial noises or patches in image classification or object detection. However, these types of perturbation may not represent the actual threats in physical world. To solve the challenge, more realistic adversarial scenarios need to be studied in different applications. Some preliminary work include verification code generation~\footnote{https://github.com/littleredhat1997/captcha-adversarial-attack}, and semantically/syntactically equivalent adversarial text generation~\cite{Lei-etal19discreteText, Ribeiro-etal18semanticallyEquNLP}.
Meanwhile, model developers need to be consistently alert to new types of attack that utilizes interpretation as the back door. For examples, it is possible to build models that predict correctly on normal data, but make mistakes on input with certain secret attacker-chosen property~\cite{Gu-etal18badnets}. Also, recently researchers found that it is possible to break data privacy by reconstructing private data merely from gradients communicated between machines~\cite{Zhu-etal19deepLeakageGrad}.

\subsection{Improving Models with Adversarial \\Samples}

The value of adversarial samples goes beyond simply serving as prewarning of model vulnerability. It is possible that the vulnerability to adversarial samples reflects some deeper generalization issues of deep models~\cite{barbu2019objectnet, Brown-etal18unrestrictedAdv}. Some preliminary work has been conducted to understand the difference between a robust model and a non-robust one. For examples, it has been shown that adversarially trained models possess better interpretability~\cite{Zhang-Zhu19interpretAdvTrained} and representations with higher quality~\cite{Tsipras-etal19odds, Santurkar-etal19synthesisSingleClf}. \cite{du2019learning} also tries to connect adversarial robustness with model credibility, where credibility measures the degree that a model's reasoning conforms with human common sense. Another challenging problem is how to properly use adversarial samples to benefit model performance, since many existing work report that training with adversarial samples will lead to performance degradation especially on large data~\cite{Kurakin-etal17atScale, Xie-etal18denoising}. Recently, ~\cite{Xie-etal19domainAdapt} shows that, by separately considering the distributions of normal data and adversarial data with batch normalization, adversarial training can be used to improve model accuracy. 

\section{Conclusion}
In this paper, we review the recent work of adversarial attack and defense by combining them with the recent advances of interpretable machine learning. Specifically, we categorize interpretation techniques into feature-level interpretation and model-level interpretation. Within each category, we investigated how the interpretation could be used for initiating adversarial attacks or designing defense approaches. After that, we briefly discuss other relations between interpretation and adversarial samples or robustness. Finally, we discuss current challenges of developing transparent and robust models, as well as potential directions to further utilizing adversarial samples.

\bibliographystyle{unsrt}
\bibliography{survey_adv_intp}

\begin{thebibliography}{100}

\bibitem{Good-etal15explaining}
Ian~J Goodfellow, Jonathon Shlens, and Christian Szegedy.
\newblock Explaining and harnessing adversarial examples.
\newblock {\em arXiv preprint arXiv:1412.6572}, 2014.

\bibitem{Szeg-etal13intriguing}
Christian Szegedy, Wojciech Zaremba, Ilya Sutskever, Joan Bruna, Dumitru Erhan,
  Ian Goodfellow, and Rob Fergus.
\newblock Intriguing properties of neural networks.
\newblock {\em arXiv preprint arXiv:1312.6199}, 2013.

\bibitem{Kurakin-etal17atScale}
Alexey Kurakin, Ian Goodfellow, and Samy Bengio.
\newblock Adversarial machine learning at scale.
\newblock 2017.

\bibitem{Lei-etal19discreteText}
Qi~Lei, Lingfei Wu, Pin-Yu Chen, Alexandros~G Dimakis, Inderjit~S Dhillon, and
  Michael Witbrock.
\newblock Discrete adversarial attacks and submodular optimization with
  applications to text classification.
\newblock {\em Systems and Machine Learning (SysML)}, 2019.

\bibitem{Liu-etal18adversarial}
Ninghao Liu, Hongxia Yang, and Xia Hu.
\newblock Adversarial detection with model interpretation.
\newblock In {\em KDD}, 2018.

\bibitem{zugner2018adversarial}
Daniel Z{\"u}gner, Amir Akbarnejad, and Stephan G{\"u}nnemann.
\newblock Adversarial attacks on neural networks for graph data.
\newblock In {\em KDD}, 2018.

\bibitem{song2018physical}
Dawn Song, Kevin Eykholt, Ivan Evtimov, Earlence Fernandes, Bo~Li, Amir
  Rahmati, Florian Tramer, Atul Prakash, and Tadayoshi Kohno.
\newblock Physical adversarial examples for object detectors.
\newblock In {\em 12th $\{$USENIX$\}$ Workshop on Offensive Technologies
  ($\{$WOOT$\}$ 18)}, 2018.

\bibitem{zeager2017adversarial}
Mary~Frances Zeager, Aksheetha Sridhar, Nathan Fogal, Stephen Adams, Donald~E
  Brown, and Peter~A Beling.
\newblock Adversarial learning in credit card fraud detection.
\newblock In {\em 2017 Systems and Information Engineering Design Symposium
  (SIEDS)}, 2017.

\bibitem{Pape-etal16practical}
Nicolas Papernot, Patrick McDaniel, Ian Goodfellow, Somesh Jha, Z~Berkay Celik,
  and Ananthram Swami.
\newblock Practical black-box attacks against machine learning.
\newblock In {\em Proceedings of the 2017 ACM on Asia conference on computer
  and communications security}, 2017.

\bibitem{Pape-etal16transferability}
Nicolas Papernot, Patrick McDaniel, and Ian Goodfellow.
\newblock Transferability in machine learning: from phenomena to black-box
  attacks using adversarial samples.
\newblock {\em arXiv preprint arXiv:1605.07277}, 2016.

\bibitem{Moosavi-etal17universalPerturb}
Seyed-Mohsen Moosavi-Dezfooli, Alhussein Fawzi, Omar Fawzi, and Pascal
  Frossard.
\newblock Universal adversarial perturbations.
\newblock In {\em CVPR}, 2017.

\bibitem{Mopuri-etal17universalDataIndependent}
Konda~Reddy Mopuri, Aditya Ganeshan, and Venkatesh~Babu Radhakrishnan.
\newblock Generalizable data-free objective for crafting universal adversarial
  perturbations.
\newblock {\em IEEE transactions on pattern analysis and machine intelligence}.

\bibitem{Thys-Ranst19surveilanceCamera}
Simen Thys, Wiebe Van~Ranst, and Toon Goedem{\'e}.
\newblock Fooling automated surveillance cameras: adversarial patches to attack
  person detection.
\newblock In {\em CVPR Workshops}, 2019.

\bibitem{Kurakin-etal16physical}
Alexey Kurakin, Ian Goodfellow, and Samy Bengio.
\newblock Adversarial examples in the physical world.
\newblock {\em arXiv preprint arXiv:1607.02533}, 2016.

\bibitem{Doshi-Kim17rigorous}
Finale Doshi-Velez and Been Kim.
\newblock Towards a rigorous science of interpretable machine learning.
\newblock {\em arXiv preprint arXiv:1702.08608}, 2017.

\bibitem{lombrozo2006structure}
Tania Lombrozo.
\newblock The structure and function of explanations.
\newblock {\em Trends in cognitive sciences}, 2006.

\bibitem{keil2006explanation}
Frank~C Keil.
\newblock Explanation and understanding.
\newblock {\em Annu. Rev. Psychol.}, pages 227--254, 2006.

\bibitem{hempel1948studies}
Carl~G Hempel and Paul Oppenheim.
\newblock Studies in the logic of explanation.
\newblock {\em Philosophy of science}, 1948.

\bibitem{Montavon-etal18methodsSurvey}
Gr{\'e}goire Montavon, Wojciech Samek, and Klaus-Robert M{\"u}ller.
\newblock Methods for interpreting and understanding deep neural networks.
\newblock {\em Digital Signal Processing}, 2018.

\bibitem{Metzen-etal17detectPerturbation}
Jan~Hendrik Metzen, Tim Genewein, Volker Fischer, and Bastian Bischoff.
\newblock On detecting adversarial perturbations.
\newblock {\em ICLR}, 2017.

\bibitem{Xie-etal18randomization}
Cihang Xie, Jianyu Wang, Zhishuai Zhang, Zhou Ren, and Alan Yuille.
\newblock Mitigating adversarial effects through randomization.
\newblock {\em arXiv preprint arXiv:1711.01991}, 2017.

\bibitem{Liao-etal18NIPSChallenge}
Fangzhou Liao, Ming Liang, Yinpeng Dong, Tianyu Pang, Xiaolin Hu, and Jun Zhu.
\newblock Defense against adversarial attacks using high-level representation
  guided denoiser.
\newblock In {\em CVPR}, 2018.

\bibitem{Xu-etal17feature}
Weilin Xu, David Evans, and Yanjun Qi.
\newblock Feature squeezing: Detecting adversarial examples in deep neural
  networks.
\newblock {\em arXiv preprint arXiv:1704.01155}, 2017.

\bibitem{Xie-etal18denoising}
Cihang Xie, Yuxin Wu, Laurens van~der Maaten, Alan~L Yuille, and Kaiming He.
\newblock Feature denoising for improving adversarial robustness.
\newblock In {\em CVPR}, 2019.

\bibitem{Madry-etalICLR18deepResistant}
Aleksander Madry, Aleksandar Makelov, Ludwig Schmidt, Dimitris Tsipras, and
  Adrian Vladu.
\newblock Towards deep learning models resistant to adversarial attacks.
\newblock {\em arXiv preprint arXiv:1706.06083}, 2017.

\bibitem{Pape-etal16distillation}
Nicolas Papernot, Patrick McDaniel, Xi~Wu, Somesh Jha, and Ananthram Swami.
\newblock Distillation as a defense to adversarial perturbations against deep
  neural networks.
\newblock In {\em 2016 IEEE Symposium on Security and Privacy (SP)}. IEEE.

\bibitem{Lu-etal17detectingRejecting}
Jiajun Lu, Theerasit Issaranon, and David Forsyth.
\newblock Safetynet: Detecting and rejecting adversarial examples robustly.
\newblock In {\em ICCV}, 2017.

\bibitem{Tao-etalNIPS18attackMeetsInterp}
Guanhong Tao, Shiqing Ma, Yingqi Liu, and Xiangyu Zhang.
\newblock Attacks meet interpretability: Attribute-steered detection of
  adversarial samples.
\newblock In {\em NIPS}, 2018.

\bibitem{Gong-etal17notTwins}
Zhitao Gong, Wenlu Wang, and Wei-Shinn Ku.
\newblock Adversarial and clean data are not twins.
\newblock {\em arXiv preprint arXiv:1704.04960}, 2017.

\bibitem{Meng-Chen17magnet}
Dongyu Meng and Hao Chen.
\newblock Magnet: a two-pronged defense against adversarial examples.
\newblock In {\em Proceedings of the 2017 ACM SIGSAC Conference on Computer and
  Communications Security}, 2017.

\bibitem{Grosse-etal17statisticDetection}
Kathrin Grosse, Praveen Manoharan, Nicolas Papernot, Michael Backes, and
  Patrick McDaniel.
\newblock On the (statistical) detection of adversarial examples.
\newblock {\em arXiv preprint arXiv:1702.06280}, 2017.

\bibitem{du2019attribution}
Mengnan Du, Ninghao Liu, Fan Yang, Shuiwang Ji, and Xia Hu.
\newblock On attribution of recurrent neural network predictions via additive
  decomposition.
\newblock In {\em The World Wide Web Conference}, 2019.

\bibitem{Zhou-etal16CAM}
Bolei Zhou, Aditya Khosla, Agata Lapedriza, Aude Oliva, and Antonio Torralba.
\newblock Learning deep features for discriminative localization.
\newblock In {\em Proceedings of the IEEE conference on computer vision and
  pattern recognition}, 2016.

\bibitem{Simonyan-etal13deepInsideCNNsaliency}
Karen Simonyan, Andrea Vedaldi, and Andrew Zisserman.
\newblock Deep inside convolutional networks: Visualising image classification
  models and saliency maps.
\newblock {\em arXiv preprint arXiv:1312.6034}, 2013.

\bibitem{Rudin19pleaseStop}
Cynthia Rudin.
\newblock Stop explaining black box machine learning models for high stakes
  decisions and use interpretable models instead.
\newblock {\em Nature Machine Intelligence}, 2019.

\bibitem{Murdoch-etal19defMethApp}
W~James Murdoch, Chandan Singh, Karl Kumbier, Reza Abbasi-Asl, and Bin Yu.
\newblock Interpretable machine learning: definitions, methods, and
  applications.
\newblock {\em arXiv preprint arXiv:1901.04592}, 2019.

\bibitem{Smilkov-etal18smoothgrad}
Daniel Smilkov, Nikhil Thorat, Been Kim, Fernanda Vi{\'e}gas, and Martin
  Wattenberg.
\newblock Smoothgrad: removing noise by adding noise.
\newblock {\em arXiv preprint arXiv:1706.03825}, 2017.

\bibitem{Tramer-etal18ensembleAdvTraining}
Florian Tram{\`e}r, Alexey Kurakin, Nicolas Papernot, Ian Goodfellow, Dan
  Boneh, and Patrick McDaniel.
\newblock Ensemble adversarial training: Attacks and defenses.
\newblock {\em arXiv preprint arXiv:1705.07204}, 2017.

\bibitem{Cao-Gong17regionClassification}
Xiaoyu Cao and Neil~Zhenqiang Gong.
\newblock Mitigating evasion attacks to deep neural networks via region-based
  classification.
\newblock In {\em ACSAC}, 2017.

\bibitem{Sundararajan-etal16integratedGradient}
Mukund Sundararajan, Ankur Taly, and Qiqi Yan.
\newblock Axiomatic attribution for deep networks.
\newblock In {\em ICML}, 2017.

\bibitem{Che-etal15distilling}
Zhengping Che, Sanjay Purushotham, Robinder Khemani, and Yan Liu.
\newblock Distilling knowledge from deep networks with applications to
  healthcare domain.
\newblock {\em arXiv preprint arXiv:1512.03542}, 2015.

\bibitem{gao2017interpretable}
Jun Gao, Ninghao Liu, Mark Lawley, and Xia Hu.
\newblock An interpretable classification framework for information extraction
  from online healthcare forums.
\newblock {\em Journal of healthcare engineering}, 2017.

\bibitem{Ribe-etal16why}
Marco~Tulio Ribeiro, Sameer Singh, and Carlos Guestrin.
\newblock Why should i trust you?: Explaining the predictions of any
  classifier.
\newblock In {\em KDD}, 2016.

\bibitem{Guo-etal18lemna}
Wenbo Guo, Dongliang Mu, Jun Xu, Purui Su, Gang Wang, and Xinyu Xing.
\newblock Lemna: Explaining deep learning based security applications.
\newblock In {\em CCS}, 2018.

\bibitem{Bigg-etal13evasion}
Battista Biggio, Igino Corona, Davide Maiorca, Blaine Nelson, Nedim
  {\v{S}}rndi{\'c}, Pavel Laskov, Giorgio Giacinto, and Fabio Roli.
\newblock Evasion attacks against machine learning at test time.
\newblock In {\em Joint European conference on machine learning and knowledge
  discovery in databases}, 2013.

\bibitem{Athalye-etal18obfuscatedGradientFalse}
Anish Athalye, Nicholas Carlini, and David Wagner.
\newblock Obfuscated gradients give a false sense of security: Circumventing
  defenses to adversarial examples.
\newblock 2018.

\bibitem{Boloor-etal19physicalAutoDrive}
Adith Boloor, Xin He, Christopher Gill, Yevgeniy Vorobeychik, and Xuan Zhang.
\newblock Simple physical adversarial examples against end-to-end autonomous
  driving models.
\newblock In {\em ICESS}. IEEE, 2019.

\bibitem{Dhurandhar-etalNIPS18constrastive}
Amit Dhurandhar, Pin-Yu Chen, Ronny Luss, Chun-Chen Tu, Paishun Ting,
  Karthikeyan Shanmugam, and Payel Das.
\newblock Explanations based on the missing: Towards contrastive explanations
  with pertinent negatives.
\newblock In {\em NIPS}, 2018.

\bibitem{Sato-etal18interpretableAdvText}
Motoki Sato, Jun Suzuki, Hiroyuki Shindo, and Yuji Matsumoto.
\newblock Interpretable adversarial perturbation in input embedding space for
  text.
\newblock In {\em IJCAI}, 2018.

\bibitem{He-etal17ensembleNotStrong}
Warren He, James Wei, Xinyun Chen, Nicholas Carlini, and Dawn Song.
\newblock Adversarial example defense: Ensembles of weak defenses are not
  strong.
\newblock In {\em 11th $\{$USENIX$\}$ Workshop on Offensive Technologies
  ($\{$WOOT$\}$ 17)}, 2017.

\bibitem{0NIPS2017}
Alexey Kurakin, Ian Goodfellow, Samy Bengio, Yinpeng Dong, Fangzhou Liao, Ming
  Liang, Tianyu Pang, Jun Zhu, Xiaolin Hu, Cihang Xie, et~al.
\newblock Adversarial attacks and defences competition.
\newblock Springer, 2018.

\bibitem{Schmidt-etal18moredata}
Ludwig Schmidt, Shibani Santurkar, Dimitris Tsipras, Kunal Talwar, and
  Aleksander Madry.
\newblock Adversarially robust generalization requires more data.
\newblock In {\em NIPS}, 2018.

\bibitem{Zhang-etal19theoretically}
Hongyang Zhang, Yaodong Yu, Jiantao Jiao, Eric Xing, Laurent~El Ghaoui, and
  Michael Jordan.
\newblock Theoretically principled trade-off between robustness and accuracy.
\newblock In {\em ICML}, 2019.

\bibitem{Cullina-etal18paclearning}
Daniel Cullina, Arjun~Nitin Bhagoji, and Prateek Mittal.
\newblock Pac-learning in the presence of adversaries.
\newblock In {\em NIPS}, 2018.

\bibitem{baker2018deep}
Nicholas Baker, Hongjing Lu, Gennady Erlikhman, and Philip~J Kellman.
\newblock Deep convolutional networks do not classify based on global object
  shape.
\newblock {\em PLoS computational biology}, 2018.

\bibitem{Shi-etal20infodrop}
Baifeng Shi, Dinghuai Zhang, Qi~Dai, Zhanxing Zhu, Yadong Mu, and Jingdong
  Wang.
\newblock Informative dropout for robust representation learning: A shape-bias
  perspective.
\newblock 2020.

\bibitem{Fidel-etal19detectSHAP}
Gil Fidel, Ron Bitton, and Asaf Shabtai.
\newblock When explainability meets adversarial learning: Detecting adversarial
  examples using shap signatures.
\newblock {\em arXiv preprint arXiv:1909.03418}, 2019.

\bibitem{Fong-Vedaldi17perturbation}
Ruth~C Fong and Andrea Vedaldi.
\newblock Interpretable explanations of black boxes by meaningful perturbation.
\newblock In {\em ICCV}, 2017.

\bibitem{Yang-etal19detectAttribution}
Puyudi Yang, Jianbo Chen, Cho-Jui Hsieh, Jane-Ling Wang, and Michael~I Jordan.
\newblock Ml-loo: Detecting adversarial examples with feature attribution.
\newblock {\em arXiv preprint arXiv:1906.03499}, 2019.

\bibitem{Zhang-etal18detectSaliency}
Chiliang Zhang, Zuochang Ye, Yan Wang, and Zhimou Yang.
\newblock Detecting adversarial perturbations with saliency.
\newblock In {\em 2018 IEEE 3rd International Conference on Signal and Image
  Processing (ICSIP)}, pages 271--275. IEEE, 2018.

\bibitem{Wang-etal20interpSafety}
Jingyuan Wang, Yufan Wu, Mingxuan Li, Xin Lin, Junjie Wu, and Chao Li.
\newblock Interpretability is a kind of safety: An interpreter-based ensemble
  for adversary defense.
\newblock In {\em KDD}, 2020.

\bibitem{Dong-etal17improveInterpretWithAdversarial}
Yinpeng Dong, Hang Su, Jun Zhu, and Fan Bao.
\newblock Towards interpretable deep neural networks by leveraging adversarial
  examples.
\newblock {\em arXiv preprint arXiv:1708.05493}, 2017.

\bibitem{Zhang-Zhu19interpretAdvTrained}
Tianyuan Zhang and Zhanxing Zhu.
\newblock Interpreting adversarially trained convolutional neural networks.
\newblock In {\em International Conference on Machine Learning}, pages
  7502--7511, 2019.

\bibitem{Ghorbani-etal19fragile}
Amirata Ghorbani, Abubakar Abid, and James Zou.
\newblock Interpretation of neural networks is fragile.
\newblock In {\em AAAI}, 2019.

\bibitem{Subramanya-etal19foolingIntp}
Akshayvarun Subramanya, Vipin Pillai, and Hamed Pirsiavash.
\newblock Fooling network interpretation in image classification.
\newblock In {\em ICCV}, 2019.

\bibitem{ZhangQ-etal18cnnTrees}
Quanshi Zhang, Yu~Yang, Haotian Ma, and Ying~Nian Wu.
\newblock Interpreting cnns via decision trees.
\newblock In {\em CVPR}, 2019.

\bibitem{Kim-etal18concepts}
Been Kim, Martin Wattenberg, Justin Gilmer, Carrie Cai, James Wexler, Fernanda
  Viegas, et~al.
\newblock Interpretability beyond feature attribution: Quantitative testing
  with concept activation vectors (tcav).
\newblock In {\em ICML}, 2018.

\bibitem{Zhou-etal18visualBasis}
Bolei Zhou, Yiyou Sun, David Bau, and Antonio Torralba.
\newblock Interpretable basis decomposition for visual explanation.
\newblock In {\em ECCV}, 2018.

\bibitem{liu2018interpretation}
Ninghao Liu, Xiao Huang, Jundong Li, and Xia Hu.
\newblock On interpretation of network embedding via taxonomy induction.
\newblock In {\em KDD}, 2018.

\bibitem{Erhan-etal09visualizingHigherLayers}
Dumitru Erhan, Yoshua Bengio, Aaron Courville, and Pascal Vincent.
\newblock Visualizing higher-layer features of a deep network.
\newblock {\em University of Montreal}, page~1.

\bibitem{olah2017feature}
Chris Olah, Alexander Mordvintsev, and Ludwig Schubert.
\newblock Feature visualization.
\newblock {\em Distill}, 2(11):e7, 2017.

\bibitem{Brown-etal18advPatch}
Tom~B Brown, Dandelion Man{\'e}, Aurko Roy, Mart{\'\i}n Abadi, and Justin
  Gilmer.
\newblock Adversarial patch.
\newblock {\em arXiv preprint arXiv:1712.09665}, 2017.

\bibitem{Sharif-etal16glasses}
Mahmood Sharif, Sruti Bhagavatula, Lujo Bauer, and Michael~K Reiter.
\newblock Accessorize to a crime: Real and stealthy attacks on state-of-the-art
  face recognition.
\newblock In {\em CCS}, 2016.

\bibitem{bengio2013representation}
Yoshua Bengio, Aaron Courville, and Pascal Vincent.
\newblock Representation learning: A review and new perspectives.
\newblock {\em IEEE transactions on pattern analysis and machine intelligence},
  2013.

\bibitem{Young-etal18survey}
Tom Young, Devamanyu Hazarika, Soujanya Poria, and Erik Cambria.
\newblock Recent trends in deep learning based natural language processing.
\newblock {\em IEEE Computational intelligenCe magazine}, 2018.

\bibitem{Hamilton-etal18survey}
William~L Hamilton, Rex Ying, and Jure Leskovec.
\newblock Representation learning on graphs: Methods and applications.
\newblock {\em arXiv preprint arXiv:1709.05584}, 2017.

\bibitem{Higgins-etal17betaVAE}
Irina Higgins, Loic Matthey, Arka Pal, Christopher Burgess, Xavier Glorot,
  Matthew Botvinick, Shakir Mohamed, and Alexander Lerchner.
\newblock beta-vae: Learning basic visual concepts with a constrained
  variational framework.
\newblock {\em ICLR}, 2017.

\bibitem{Panigrahi-etal19word2sense}
Abhishek Panigrahi, Harsha~Vardhan Simhadri, and Chiranjib Bhattacharyya.
\newblock Word2sense: Sparse interpretable word embeddings.
\newblock In {\em ACL}, 2019.

\bibitem{liu2019single}
Ninghao Liu, Qiaoyu Tan, Yuening Li, Hongxia Yang, Jingren Zhou, and Xia Hu.
\newblock Is a single vector enough? exploring node polysemy for network
  embedding.
\newblock In {\em KDD}, 2019.

\bibitem{Ma-etal19disentangled}
Jianxin Ma, Chang Zhou, Peng Cui, Hongxia Yang, and Wenwu Zhu.
\newblock Learning disentangled representations for recommendation.
\newblock In {\em Advances in Neural Information Processing Systems}, 2019.

\bibitem{Ghorbani-etal19automaticConcept}
Amirata Ghorbani, James Wexler, James~Y Zou, and Been Kim.
\newblock Towards automatic concept-based explanations.
\newblock In {\em NeurIPS}, 2019.

\bibitem{Mathew-etal19polarInterpWordEmbd}
Binny Mathew, Sandipan Sikdar, Florian Lemmerich, and Markus Strohmaier.
\newblock The polar framework: Polar opposites enable interpretability of
  pre-trained word embeddings.
\newblock In {\em The World Wide Web Conference}, 2020.

\bibitem{Kim-etal14thebayesian}
Been Kim, Cynthia Rudin, and Julie~A Shah.
\newblock The bayesian case model: A generative approach for case-based
  reasoning and prototype classification.
\newblock In {\em NIPS}, 2014.

\bibitem{Koh-Liang17influencefunction}
Pang~Wei Koh and Percy Liang.
\newblock Understanding black-box predictions via influence functions.
\newblock In {\em ICML}, 2017.

\bibitem{Velickovic-etal18graphAttention}
Petar Veli{\v{c}}kovi{\'c}, Guillem Cucurull, Arantxa Casanova, Adriana Romero,
  Pietro Lio, and Yoshua Bengio.
\newblock Graph attention networks.
\newblock {\em arXiv preprint arXiv:1710.10903}, 2017.

\bibitem{Chen-etal19thisLooksthat}
Chaofan Chen, Oscar Li, Daniel Tao, Alina Barnett, Cynthia Rudin, and
  Jonathan~K Su.
\newblock This looks like that: deep learning for interpretable image
  recognition.
\newblock In {\em NeurIPS}, 2019.

\bibitem{Dai-etal19advTrainingGraph}
Quanyu Dai, Xiao Shen, Liang Zhang, Qiang Li, and Dan Wang.
\newblock Adversarial training methods for network embedding.
\newblock In {\em The World Wide Web Conference}, 2019.

\bibitem{Ilyas-etal19notBugs}
Andrew Ilyas, Shibani Santurkar, Dimitris Tsipras, Logan Engstrom, Brandon
  Tran, and Aleksander Madry.
\newblock Adversarial examples are not bugs, they are features.
\newblock {\em arXiv preprint arXiv:1905.02175}, 2019.

\bibitem{Wang-etal20highFreq}
Haohan Wang, Xindi Wu, Zeyi Huang, and Eric~P Xing.
\newblock High-frequency component helps explain the generalization of
  convolutional neural networks.
\newblock In {\em CVPR}, 2020.

\bibitem{Lundberg-etal17SHAP}
Scott~M Lundberg and Su-In Lee.
\newblock A unified approach to interpreting model predictions.
\newblock In {\em NIPS}, 2017.

\bibitem{Li-etal17detectConvFilterStat}
Xin Li and Fuxin Li.
\newblock Adversarial examples detection in deep networks with convolutional
  filter statistics.
\newblock In {\em ICCV}, 2017.

\bibitem{Feinman-etal17detectArtifacts}
Reuben Feinman, Ryan~R Curtin, Saurabh Shintre, and Andrew~B Gardner.
\newblock Detecting adversarial samples from artifacts.
\newblock {\em arXiv preprint arXiv:1703.00410}, 2017.

\bibitem{Tsipras-etal19odds}
Dimitris Tsipras, Shibani Santurkar, Logan Engstrom, Alexander Turner, and
  Aleksander Madry.
\newblock Robustness may be at odds with accuracy.
\newblock {\em arXiv preprint arXiv:1805.12152}, 2018.

\bibitem{Santurkar-etal19synthesisSingleClf}
Shibani Santurkar, Andrew Ilyas, Dimitris Tsipras, Logan Engstrom, Brandon
  Tran, and Aleksander Madry.
\newblock Image synthesis with a single (robust) classifier.
\newblock In {\em NeurIPS}, 2019.

\bibitem{Ross-Velez18regularizeGradient}
Ninghao Liu, Xiao Huang, Jundong Li, and Xia Hu.
\newblock On interpretation of network embedding via taxonomy induction.
\newblock In {\em KDD}, 2018.

\bibitem{Zhu-Li20purification}
Zeyuan Allen-Zhu and Yuanzhi Li.
\newblock Feature purification: How adversarial training performs robust deep
  learning.
\newblock {\em arXiv preprint arXiv:2005.10190}, 2020.

\bibitem{guo2020survey}
Ruocheng Guo, Lu~Cheng, Jundong Li, P~Richard Hahn, and Huan Liu.
\newblock A survey of learning causality with data: Problems and methods.
\newblock {\em ACM Computing Surveys (CSUR)}, 2020.

\bibitem{moraffah2020causal}
Raha Moraffah, Mansooreh Karami, Ruocheng Guo, Adrienne Raglin, and Huan Liu.
\newblock Causal interpretability for machine learning-problems, methods and
  evaluation.
\newblock {\em ACM SIGKDD Explorations Newsletter}, 2020.

\bibitem{wachter2017counterfactual}
Sandra Wachter, Brent Mittelstadt, and Chris Russell.
\newblock Counterfactual explanations without opening the black box: Automated
  decisions and the gdpr.
\newblock {\em Harv. JL \& Tech.}, 31:841, 2017.

\bibitem{Li-Jurafsky17erasure}
Jiwei Li, Will Monroe, and Dan Jurafsky.
\newblock Understanding neural networks through representation erasure.
\newblock {\em arXiv preprint arXiv:1612.08220}, 2016.

\bibitem{Dabkowski-Gal17realtimeSaliency}
Piotr Dabkowski and Yarin Gal.
\newblock Real time image saliency for black box classifiers.
\newblock In {\em NIPS}, 2017.

\bibitem{ronneberger2015u}
Olaf Ronneberger, Philipp Fischer, and Thomas Brox.
\newblock U-net: Convolutional networks for biomedical image segmentation.
\newblock In {\em International Conference on Medical image computing and
  computer-assisted intervention}. Springer, 2015.

\bibitem{Schulz-etal19restrictingFlow}
Karl Schulz, Leon Sixt, Federico Tombari, and Tim Landgraf.
\newblock Restricting the flow: Information bottlenecks for attribution.
\newblock In {\em ICLR}, 2019.

\bibitem{narendra2018explaining}
Tanmayee Narendra, Anush Sankaran, Deepak Vijaykeerthy, and Senthil Mani.
\newblock Explaining deep learning models using causal inference.
\newblock {\em arXiv preprint arXiv:1811.04376}, 2018.

\bibitem{Harraton-etal18causalExplain}
Michael Harradon, Jeff Druce, and Brian Ruttenberg.
\newblock Causal learning and explanation of deep neural networks via
  autoencoded activations.
\newblock {\em arXiv preprint arXiv:1802.00541}, 2018.

\bibitem{liu2020interpretations}
Ninghao Liu, Yunsong Meng, Xia Hu, Tie Wang, and Bo~Long.
\newblock Are interpretations fairly evaluated? a definition driven pipeline
  for post-hoc interpretability.
\newblock {\em arXiv preprint arXiv:2009.07494}, 2020.

\bibitem{lim2009and}
Brian~Y Lim, Anind~K Dey, and Daniel Avrahami.
\newblock Why and why not explanations improve the intelligibility of
  context-aware intelligent systems.
\newblock In {\em Proceedings of the SIGCHI Conference on Human Factors in
  Computing Systems}. ACM, 2009.

\bibitem{narayanan2018humans}
Menaka Narayanan, Emily Chen, Jeffrey He, Been Kim, Sam Gershman, and Finale
  Doshi-Velez.
\newblock How do humans understand explanations from machine learning systems?
  an evaluation of the human-interpretability of explanation.
\newblock {\em arXiv preprint arXiv:1802.00682}, 2018.

\bibitem{yang2019xfake}
Fan Yang, Shiva~K Pentyala, Sina Mohseni, Mengnan Du, Hao Yuan, Rhema Linder,
  Eric~D Ragan, Shuiwang Ji, and Xia~Ben Hu.
\newblock Xfake: Explainable fake news detector with visualizations.
\newblock In {\em WWW}, 2019.

\bibitem{bichindaritz2006case}
Isabelle Bichindaritz and Cindy Marling.
\newblock Case-based reasoning in the health sciences: What's next?
\newblock {\em Artificial intelligence in medicine}, 2006.

\bibitem{Kim-etal16examples}
Been Kim, Rajiv Khanna, and Oluwasanmi~O Koyejo.
\newblock Examples are not enough, learn to criticize! criticism for
  interpretability.
\newblock In {\em NIPS}, 2016.

\bibitem{Goyal-etal19counterfactual}
Yash Goyal, Ziyan Wu, Jan Ernst, Dhruv Batra, Devi Parikh, and Stefan Lee.
\newblock Counterfactual visual explanations.
\newblock In {\em ICML}, 2019.

\bibitem{zhao2019personalized}
Guoshuai Zhao, Hao Fu, Ruihua Song, Tetsuya Sakai, Zhongxia Chen, Xing Xie, and
  Xueming Qian.
\newblock Personalized reason generation for explainable song recommendation.
\newblock {\em ACM Transactions on Intelligent Systems and Technology (TIST)},
  2019.

\bibitem{Heo-etal19foolInterp}
Juyeon Heo, Sunghwan Joo, and Taesup Moon.
\newblock Fooling neural network interpretations via adversarial model
  manipulation.
\newblock {\em arXiv preprint arXiv:1902.02041}, 2019.

\bibitem{slack2020fooling}
Dylan Slack, Sophie Hilgard, Emily Jia, Sameer Singh, and Himabindu Lakkaraju.
\newblock Fooling lime and shap: Adversarial attacks on post hoc explanation
  methods.
\newblock In {\em AAAI}, 2020.

\bibitem{Dombrowski-etal19manipulate}
Ann-Kathrin Dombrowski, Maximillian Alber, Christopher Anders, Marcel
  Ackermann, Klaus-Robert M{\"u}ller, and Pan Kessel.
\newblock Explanations can be manipulated and geometry is to blame.
\newblock In {\em NeurIPS}, 2019.

\bibitem{Levine-etal19certifyRobustInterp}
Alexander Levine, Sahil Singla, and Soheil Feizi.
\newblock Certifiably robust interpretation in deep learning.
\newblock {\em arXiv preprint arXiv:1905.12105}, 2019.

\bibitem{Battaglia-etal18google}
Peter~W Battaglia, Jessica~B Hamrick, Victor Bapst, Alvaro Sanchez-Gonzalez,
  Vinicius Zambaldi, Mateusz Malinowski, Andrea Tacchetti, David Raposo, Adam
  Santoro, Ryan Faulkner, et~al.
\newblock Relational inductive biases, deep learning, and graph networks.
\newblock {\em arXiv preprint arXiv:1806.01261}, 2018.

\bibitem{Sabour-Hinton17routing}
Sara Sabour, Nicholas Frosst, and Geoffrey~E Hinton.
\newblock Dynamic routing between capsules.
\newblock In {\em NIPS}, 2017.

\bibitem{Peters-etal17elements}
Jonas Peters, Dominik Janzing, and Bernhard Sch{\"o}lkopf.
\newblock {\em Elements of causal inference: foundations and learning
  algorithms}.
\newblock MIT press, 2017.

\bibitem{Ribeiro-etal18semanticallyEquNLP}
Marco~Tulio Ribeiro, Sameer Singh, and Carlos Guestrin.
\newblock Semantically equivalent adversarial rules for debugging nlp models.
\newblock In {\em Proceedings of the 56th Annual Meeting of the Association for
  Computational Linguistics (Volume 1: Long Papers)}, pages 856--865, 2018.

\bibitem{Gu-etal18badnets}
Tianyu Gu, Brendan Dolan-Gavitt, and Siddharth Garg.
\newblock Badnets: Identifying vulnerabilities in the machine learning model
  supply chain.
\newblock {\em arXiv preprint arXiv:1708.06733}, 2017.

\bibitem{Zhu-etal19deepLeakageGrad}
Ligeng Zhu, Zhijian Liu, and Song Han.
\newblock Deep leakage from gradients.
\newblock {\em arXiv preprint arXiv:1906.08935}, 2019.

\bibitem{barbu2019objectnet}
Andrei Barbu, David Mayo, Julian Alverio, William Luo, Christopher Wang, Dan
  Gutfreund, Josh Tenenbaum, and Boris Katz.
\newblock Objectnet: A large-scale bias-controlled dataset for pushing the
  limits of object recognition models.
\newblock In {\em NeurIPS}, 2019.

\bibitem{Brown-etal18unrestrictedAdv}
Tom~B Brown, Nicholas Carlini, Chiyuan Zhang, Catherine Olsson, Paul
  Christiano, and Ian Goodfellow.
\newblock Unrestricted adversarial examples.
\newblock {\em arXiv preprint arXiv:1809.08352}, 2018.

\bibitem{du2019learning}
Mengnan Du, Ninghao Liu, Fan Yang, and Xia Hu.
\newblock Learning credible deep neural networks with rationale regularization.
\newblock In {\em ICDM}, 2019.

\bibitem{Xie-etal19domainAdapt}
Cihang Xie, Mingxing Tan, Boqing Gong, Jiang Wang, Alan Yuille, and Quoc~V Le.
\newblock Adversarial examples improve image recognition.
\newblock {\em arXiv preprint arXiv:1911.09665}, 2019.

\end{thebibliography}

\end{document}